\documentclass{article}

\PassOptionsToPackage{numbers}{natbib}
\usepackage[preprint]{neurips_data_2024}

\usepackage[utf8]{inputenc} % allow utf-8 input
\usepackage[T1]{fontenc}    % use 8-bit T1 fonts
\usepackage{hyperref}       % hyperlinks
\usepackage{url}            % simple URL typesetting
\usepackage{booktabs}       % professional-quality tables
\usepackage{amsfonts}       % blackboard math symbols
\usepackage{nicefrac}       % compact symbols for 1/2, etc.
\usepackage{microtype}      % microtypography
\usepackage{xcolor}         % colors
\usepackage{float}
\usepackage{multirow}
\usepackage{graphicx}
\usepackage{cleveref}
\usepackage{subfigure}
\usepackage{subcaption}
\usepackage{rotating}

\title{Bag of Tricks for Multimodal AutoML \\ with Image, Text, and Tabular Data}

\author{%
  Zhiqiang Tang \\
  Amazon Web Services\\
  \texttt{zqtang@amazon.com} \\
  \And
  Zihan Zhong \thanks{Work done while interning at Amazon Web Services.} \\
  Tsinghua University \\  \texttt{zhongzh22@mails.tsinghua.edu.cn} \\
  \AND
  Tong He \\
  Amazon Web Services \\
  \texttt{htong@amazon.com} \\
  \And
  Gerald Friedland \\
  Amazon Web Services \\
  \texttt{gfriedla@amazon.com} \\
}

\begin{document}

\maketitle
\begin{abstract}
  This paper studies the best practices for automatic machine learning (AutoML). While previous AutoML efforts have predominantly focused on unimodal data, the multimodal aspect remains under-explored. Our study delves into classification and regression problems involving flexible combinations of image, text, and tabular data. We curate a benchmark comprising 22 multimodal datasets from diverse real-world applications, encompassing all 4 combinations of the 3 modalities. Across this benchmark, we scrutinize design choices related to multimodal fusion strategies, multimodal data augmentation, converting tabular data into text, cross-modal alignment,  and handling missing modalities. Through extensive experimentation and analysis, we distill a collection of effective strategies and consolidate them into a unified pipeline, achieving robust performance on diverse datasets.
\end{abstract}
\section{Introduction}
Automatic machine learning (AutoML)~\citep{ledell2020h2o, vakhrushev2021lightautoml} aims to automate the process of building ML models, including data preprocessing~\citep{cubuk2018autoaugment}, model selection~\citep{arango2023quick}, hyperparameter tuning~\citep{feurer2019hyperparameter}, model evaluation~\citep{raschka2018model}, and so on. As the diversity and complexity of state-of-the-art ML techniques grow, AutoML becomes increasingly important to democratize these techniques for broader communities with little or no ML background~\citep{janet2020machine, greener2022guide}. It also provides experienced practitioners with "hands-off-the-wheel" solutions for tackling numerous distinct problems. AutoML systems~\citep{erickson2020autogluon, feurer2022auto} typically offer user-friendly APIs and robust performance in solving real-world problems. Central to an AutoML system is a collection of "tricks" that significantly enhance performance.

Many studies~\citep{zimmer2021auto, feurer2022auto, erickson2020autogluon,ledell2020h2o} focus on the best practices of modeling tabular (categorical or numeric) data. While there has been some progress in other data modalities such as image~\citep{he2019bag} and text~\citep{joulin2016bag}, a noticeable gap exists for multimodal data. Many real-world applications~\citep{kiela2020hateful, plummer2015flickr30k} involve multiple modalities such as image, text, and tabular data. For example, \Cref{fig:multimodal-table-demo} illustrates actual data from the PetFinder Challenge~\citep{zhang2019petfinder}, which aims to predict the speed of pet adoptions. The input of this task includes pet's profile picture, text descriptions, categorial fields like gender and color, and numeric fields like age and adoption fee. This paper considers tables of this form where rows contain IID samples, and the columns are predictive features. We refer to the value in a particular row and column as a field, which can be an image path, a long text passage, a categorical or numeric value. We focus on the classification and regression tasks where each sample has a single categorical or numeric value to predict.

This paper examines a collection of tricks for multimodal supervised learning with a mix of image, text, and tabular data. Our study includes basic procedure refinements (such as greedy soup~\citep{wortsman2022model} and gradient clipping~\citep{zhang2019gradient}) for training models, as well as advanced strategies specialized for multimodal learning~\citep{ngiam2011multimodal}. Key aspects of multimodal learning include determining the best multimodal fusion strategies (early~\citep{zhang2023meta} vs. late fusion~\citep{vielzeuf2018centralnet}, parallel~\citep{vielzeuf2018centralnet} vs. sequential fusion~\citep{swamy2024multimodn}), conducting data augmentation for multimodal data (independent~\citep{muller2021trivialaugment, wei2019eda} vs. joint augmentation~\citep{liu2022learning}, input~\citep{muller2021trivialaugment, wei2019eda} vs. feature augmentation~\citep{verma2019manifold}), assessing the usefulness of cross-modal alignment~\citep{radford2021learning} for supervised learning, and handling samples with missing modalities~\citep{zeng2022tag}. For each aspect, we investigate multiple strategy alternatives. Rather than proposing new techniques, our goal is to gather existing ones from various literatures and codebases and conduct an empirical evaluation in a novel yet practical setting with flexible mixtures of image, text, and tabular modalities.

We evaluate these tricks on a new benchmark of 22 datasets collected from various domains and real-world applications~\citep{chhavi2020memotion, nakamura2019r, gao2018action}. This benchmark encompasses all 4 possible combinations of image, text, and tabular modalities. Our findings indicate that while some tricks consistently perform well across the entire benchmark, many are effective for certain modality combinations but struggle to generalize across others. Ultimately, we employ a learnable ensembling method~\citep{caruana2004ensemble} to integrate all the techniques into a single pipeline, resulting in superior performance compared to each individual technique. The weights assigned to each technique in the ensemble not only reveal their relative importance but also provide insights into the specific aspects of multimodal learning they address. Interestingly, some tricks, although not improving performance on their own, contribute positively to the ensemble, highlighting the complementary nature of different tricks.

% By examining these tricks, we aim to answer several questions: 1) Which tricks remain robustly performant across a diverse set of real-world applications? Identifying such tricks is critical for building multimodal AutoML systems. 2) Which tricks work only for specific modality compositions? For example, an effective trick for image+text data may not generalize well to image+text+tabular data. 3) Which tricks struggle to generalize across different applications even within the same modality combination? These unstable tricks are generally unsuitable for AutoML. 4) How can we achieve the best performance by incorporating all the promising tricks? Stacking two tricks together in training one model may not yield compound gains.

% To answer these questions, we evaluate the tricks on a new benchmark of datasets collected from various domains (such as healthcare, e-commerce, and social media) and real-world applications (such as skin disease classification, customer review prediction, and sentiment analysis of internet memes). The benchmark covers all four combinations of image, text, and tabular modalities. Through this benchmark, we find that basic tricks and some advanced ones, such as late fusion, input data augmentation, and modality dropout, can bring robust performance gains across the entire benchmark. The multimodal alignment trick favors the image+text combination over others. After identifying a set of promising tricks, we use a learnable ensembling method to integrate them into a single pipeline, producing better performance compared to each individual trick.

\begin{table}[t]
    \caption{Example of multimodal data with image (Image), text (Description), numeric (Age, Fee), and categorical (Color1, Gender) features from the Petfinder Challenge. The features shown here are a subset of the total 23 features. Based on these features, the task is to predict the pet adoption speed.}
    \centering
    \includegraphics[width=0.95\linewidth]{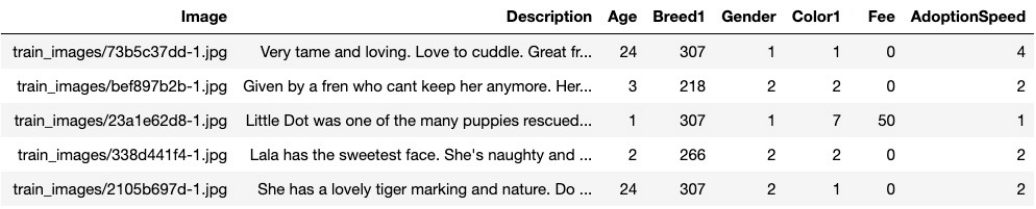}
    
    \label{fig:multimodal-table-demo}
    \vspace{-0.5em}
\end{table}

\section{Related Work}\label{sec:related_work_full}
AutoML~\citep{Karmaker2020AutoMLTD, Zhang2023AutoMLGPTAM, Tornede2023AutoMLIT} is an emerging field that aims to incorporate best practices in machine learning (ML) and automate the process of building ML models. An AutoML workflow typically involves various stages of a ML pipeline, including data preparation, feature engineering, model generation, and model evaluation. Numerous AutoML studies~\citep{ledell2020h2o, erickson2020autogluon, feurer2022auto, vakhrushev2021lightautoml} focus on tabular data. For instance, Auto-Sklearn~\citep{feurer2022auto} leverages Bayesian optimization, meta-learning, and ensemble construction to provide robust solutions for classification and regression tasks. AutoGluon-Tabular~\citep{erickson2020autogluon} uses multi-layer stacking to ensemble diverse tabular models, while H2O AutoML~\citep{ledell2020h2o} offers an end-to-end platform for automating the entire machine learning process, including automatic training and ensemble of diverse algorithms. FLAML~\citep{wang2021flaml} optimizes for low computational cost in the hyperparameter search. Beyond tabular data, there are AutoML solutions for image and text data as well. TIMM~\citep{rw2019timm}  provides a collection of techniques for image classification, and FastText~\citep{joulin2016bag} offers a simple and efficient solution for training text classifiers. In contrast to these previous efforts, our work focuses on multimodal AutoML and explores a bag of tricks to improve model performance.

% AutoML is an emerging field that aims to incorporate best ML practices and automate the process of building ML models. Conceptually, an AutoML workflow involves various stages of a machine learning pipeline including data preparation, feature engineering, model generation and model evaluation. There are many AutoML studies on tabular data. For instance, Auto-Sklearn leverages bayesian optimization, meta-learning and ensemble construction to provide a robust solution for classification and regression tasks. AutoGluon-Tabular uses multi-layer stacking to ensemble diverse tabular models. H2O AutoML provides an end-to-end platform for automating the machine learning process, including automatic training and ensemble of a diverse set of algorithms. FLAML optimizes for low computational cost in the hyperparameter search. In addition to these frameworks designed for tabular data, there are also solutions developed for image or text data. For example, TIMM provides a bag of tricks for image classification, and FastText provides a simple and efficient solution for training text classifier.  
% In contrast, we focus on multimodal AutoML and dive into a bag of tricks useful for improving model performance.

% Instead of building a new AutoML toolbox, we leverage AutoMM as the code base and dive into a bag of tricks useful for improving model performance.

% With an emphasis on practicality, 
% Recently, AutoMM emerged as a multimodal AutoML toolbox for image, text, and tabular data.

Benchmark datasets play a central role in ML research, serving as valuable resources for training and evaluating ML models. Since advancement on established benchmarks is viewed as an indicator of progress, researchers are encouraged to make design choices that maximize performance on benchmarks. Large and representative benchmarks have spurred significant progress in tabular AutoML~\citep{Gijsbers2022AMLBAA, Vanschoren2014OpenMLNS}, natural language processing~\citep{Wang2018GLUEAM, Wang2019SuperGLUEAS}, computer vision~\citep{Russakovsky2014ImageNetLS, Lin2014MicrosoftCC}.
 However, we are not aware of any analogous benchmarks for evaluating multimodal AutoML with flexible image, text, and tabular combinations. There do exist a multimodal AutoML benchmark of text+tabular datasets~\citep{shi2021benchmarking}, but it doesn’t involve image data. There are multimodal learning benchmarks with images~\citep{Fu2023MMEAC, Liu2023MMBenchIY, Yu2023MMVetEL,Li2023SEEDBenchBM}, but they are main for the generative tasks. In contrast, this paper focuses on the discriminative tasks. Moreover, our benchmark, involving all combinations of image, text, and tabular data, complements MultiBench ~\citep{liang2021multibench}, which covers more modalities but only partial modality combinations. Aside from MM-IMDB, an image+text dataset, the other datasets in MultiBench involve other modalities such as video and audio.

Along with the increasing model capacity, multimodal learning~\citep{Xu2022MultimodalLW, zhang2023meta} are attracting increasing attention with the extraordinary potential to build human-like AI agents to interact the multimodal world. Multimodal learning is a multi-disciplinary area involving domain knowledge about different modalities such as vision, language, audio, and various sensor data. Due to the complexity of multimodal learning, studies are usually conducted on specific modality combinations such as audio+vision~\citep{Lin2022VisionTA, Zhu2020DeepAL, Afouras2018DeepAS} and vision+language~\citep{Kim2021ViLTVT, Li2023BLIP2BL,Liu2023VisualIT}. In contrast, this paper explores arbitrary combinations of image, text, and tabular data. The heterogeneity of modalities bring unique challenges for multimodal learning. Prior works tried to address them from different perspectives. For example, CLIP~\citep{radford2021learning} aligns the image and text modalities in the feature space through large-scale contrastive training. LeMDA~\citep{liu2022learning} aims to learn multimodal data augmentation jointly in the feature space. Some works~\citep{ma2021smil,zhao2021missing,zeng2022tag,lee2023multimodal} also proposed methods to address the common issue of missing modalities in real-world scenarios. Among all the topics, multimodal fusion strategies~\citep{vielzeuf2018centralnet,gadzicki2020early,zhang2021deep,boulahia2021early,yao2022modality,zhang2023meta,swamy2024multimodn} receive the most attention. There exist both early fusion and late fusion models, as well sequential fusion~\citep{swamy2024multimodn}. Recently, many studies use large language models (LLMs)~\citep{Brown2020LanguageMA} as the fusion module in multimodal large language models (MLMMs)~\citep{Yin2023ASO,Han2023OneLLMOF}. This paper examines these technical designs on supervised learning with flexible image, text, and tabular combinations.

\section{A Benchmark for Supervised Learning with Image/Text/Tabular Data}
\label{sec:dataset}

\begin{table}[t]
    \caption{A multimodal benchmark including all the 4 combinations of image, text, and tabular data.}
    \large
    \centering
    \resizebox{\textwidth}{!}{%
    \begin{tabular}{lccccccccc}
        \toprule
         \textbf{Dataset ID} & \textbf{\#Train} & \textbf{\#Test} & \textbf{\#Img.} & \textbf{\#Text.} & \textbf{\#Cat.} & \textbf{\#Num.} &  \textbf{Task} & \textbf{Metric} & \textbf{Prediction Target} \\
         % \cline{2-13}
         \midrule
         fake & 12725 & 3182 & 0 & 2 & 3 & 0 &  binary &  roc-auc  & whether job postings are fake\\
         airbnb & 18316 & 4579 & 0 & 15 & 47 & 23 & multiclass & accuracy & price label of Airbnb listing \\
         channel & 20284 & 5071 & 0 & 1 & 1 & 15 & multiclass & accuracy & news category to which article belongs\\
        qaa & 4863 & 1216 & 0 & 3 & 1 & 0  &  regression & $R^2$ & subjective type of answer (in relation to question) \\
        qaq & 4863 & 1216 & 0 & 3 & 1 & 0  &  regression & $R^2$ & subjective type of question (in relation to answer)  \\
        cloth & 18788 & 4698 & 0 & 2 & 3 & 1 &  regression & $R^2$ & customer review score for clothing item \\
        
        \midrule
        ptech & 749 & 200 & 1 & 1 & 0 & 0 & binary &  roc-auc & whether the smear persuasive technique is used\\
        memotion & 6992 & 1878 & 1 & 1 & 0 & 0 & multiclass & accuracy & sentiment associated with Internet memes  \\
        food101 & 13613 & 4547 & 1 & 1 & 0 & 0 & multiclass & accuracy & food category \\
        aep & 1117 & 280 & 1 & 1 & 0 & 0 & multiclass & accuracy & action category \\
        fakeddit & 16277 & 5723 & 1 & 1 & 0 & 0 & multiclass & accuracy & fake news category \\
        
        \midrule
        ccd & 1126 & 374 & 1 & 0 & 2 & 0 & binary & roc-auc & whether ego-vehicles are involved in the accident \\
        HAM & 8010 & 2005 & 1 & 0 & 3 & 1 & multiclass & accuracy & pigmented skin lesions category \\
        wikiart & 15278 & 5084 & 1 & 0 & 2 & 0 & multiclass & accuracy & artist category \\
        cd18 & 2533 & 632 & 1 & 0 & 7 & 10 & regression & $R^2$ & price of cellphones \\
        DVM & 10403 & 3186 & 1 & 0 & 13 & 0 & regression & $R^2$ &  selling price of cars\\
        
        \midrule
        petfinder & 11994 & 2999 & 1 & 1 & 15 & 5 & multiclass &   quadratic\_kappa
 & adoption speed of pets \\
        covid  & 707 & 222 & 1 & 2 & 9 & 3 & multiclass & accuracy & pneumonia category \\
        artm & 573 & 177 & 1 & 3  & 3 & 1 & multiclass & accuracy & art movement category \\
        seattle & 3024 & 755 & 1 & 6 & 19 & 31 & regression & $R^2$ & price of living in a hotel \\
        goodreads & 11252 & 3748 & 1 & 3 & 1 & 1 & regression & $R^2$ & rating of books\\
        KARD & 15645 & 5212 & 1 & 2 & 4 & 12 & regression & $R^2$ & rating of animes \\
        \bottomrule
        
    \end{tabular}
    
    }
    \label{tab:dataset_info}
\end{table}

We introduce the first benchmark for evaluating multimodal supervised learning with flexible combinations of images, text, and tabular data. Our benchmark includes all 4 possible combinations of these 3 modalities, with at least 5 datasets for each combination, totaling 22 datasets. These datasets span a wide array of domains, including healthcare \citep{tschandl2018ham10000}, social media \citep{chhavi2020memotion}, food \citep{bossard2014food}, transportation \citep{jahangiri2015applying}, arts \citep{audry2021art}, and e-commerce \citep{haque2018sentiment}. Technically, the benchmark comprises a mix of classification and regression tasks and includes a diverse set of image, text, categorical, and numeric columns. \Cref{tab:dataset_info} demonstrates the diversity of the datasets in terms of training sample numbers (ranging from 1k to 20k), problem types (binary, multiclass, regression), number of features (0 to 47), and types of features (image+text, image+tabular, text+tabular, image+text+tabular). Some of these datasets naturally come with missing modalities (\Cref{tab:missing_dataset_info} ), reflecting real-world scenarios where some samples may have incomplete data during training or inference. For a comprehensive introduction to the datasets, refer to \Cref{sec:benchmark_details}.

\begin{table}[t]\large
    \caption{The missing modality ratios (\%) of some datasets. There are 12 datasets that naturally have missing modalities. ``Tr.'' and ``Te.'' denotes the training and test sets, respectively.}
    \centering
    \resizebox{0.9\textwidth}{!}{%
    \begin{tabular}{lcccccccccc}
        \toprule
         \textbf{Dataset ID} & \textbf{\#Train} & \textbf{\#Test} & \textbf{Tr. Img.} & \textbf{Tr. Text.} & \textbf{Tr. Cat.} & \textbf{Tr. Num.} & \textbf{Te. Img.} & \textbf{Te. Text.} & \textbf{Te. Cat.} & \textbf{Te. Num.} \\
         % \cline{2-13}
         \midrule
         fake & 12725 & 3182 & / & / &41.63&/ &/ &/ &41.67&  / \\
         airbnb & 18316 & 4579 & / & 19.83 & 5.64 & 11.11 &/  & 19.87 & 5.70 & 11.06 \\
        cloth & 18788 & 4698 & /  & 9.91 & 0.06 &/ & / &9.85 & 0.06 & / \\
        
        \midrule
        memotion & 6992 & 1878 &  /  &0.06 &  /  & /  & /  & 0.96  & /& /\\
        % food101 & 13613 & 4547 & 0.12 &   /  &  /   &  /   &  / & / & / & /  \\
        fakeddit & 16277 & 5723 & 0.04 & / & / & / & 0.11 & / & / & / \\
        \midrule

        HAM & 8010 & 2005 & /  & / & / &0.50 & / & / & / &0.80\\

        cd18 & 2533 & 632 & 0.10 & / &0.40& & / &  / & 0.02 & /\\
        DVM & 10403 & 3186 & 18.23 & / & / & / & 18.68 & / & / &/\\
        \midrule
        petfinder & 11994 & 2999 & / &0.08&0.53 & / & / &0.13&0.53 & /\\
        covid  & 707 & 222 &/ & 37.10 & 42.86 & 47.17 &/& 37.16 & 41.71 & 47.15 \\
        % artm & 573 & 177 & /  & / &0.05 & / & / & / & / & / \\
        seattle & 3024 & 755 & / & 8.29 & 5.17 & 3.86 &  / & 7.99 &5.21&4.40 \\
         KARD & 15645 & 5212 &  9.40 & 0.04& / & / &8.94 & 0.03 & / & / \\
        \bottomrule
        
    \end{tabular}}

    \label{tab:missing_dataset_info}
\end{table}

\section{Bag of Tricks for Multimodal AutoML}
\label{sec:exp}

% Based on AutoMM framework~\citep{tang2024autogluon}, we conduct thorough experiments to investigate some key tricks for multimodal learning using our benchmark. The tricks we examine include: common basic training tricks(e.g., weight decay), different multimodal fusion structures, alignment among different modalities during fine-tuning, modality missingness handling, and ensembling framework to boost the performance of the final model.
\subsection{Baseline}
\label{sec:exp_baseline}

An AutoML system is an end-to-end (from raw data to predictions) ML pipeline that includes data preprocessing, feature engineering, model selection, model training, hyperparameter optimization, and model evaluation. Building such a system involves implementing various ML techniques and integrating them into a single pipeline with careful system design. Our goal is not to develop a multimodal AutoML system from scratch but to examine a collection of tricks for training high-performance models. Given that many factors in an AutoML system can affect model performance, it is crucial to start with a solid baseline that contains all components and demonstrates reasonable performance. After comparing several AutoML toolboxes, such as Auto-Sklearn~\citep{feurer2022auto}, AutoKeras~\citep{jin2023autokeras}, FLAML~\citep{wang2021flaml}, LightAutoML~\citep{vakhrushev2021lightautoml},  AutoMM~\citep{tang2024autogluon}, TPOT~\citep{olson2016tpot}, and H2O AutoML~\citep{ledell2020h2o}, we chose AutoMM. AutoMM offers state-of-the-art performance in multimodal supervised learning and provides an open-source codebase, making it a good baseline for our experiments.

Our baseline follows the best quality presets of AutoMM, with a focus of fine-tuning pre-trained models. Specifically, it employs a late-fusion architecture, using Swin Transformer Large~\citep{liu2021swin}, DeBERTa-V3~\citep{he2021debertav3}, and FT-transformer~\citep{gorishniy2021revisiting} to independently process the image, text, and tabular data. These are followed by a multilayer perceptron (MLP)~\citep{popescu2009multilayer} for feature fusion and a linear head for prediction. The entire model is trained for 20 epochs with mixed precision~\citep{micikevicius2017mixed}, weight decay, cosine learning rate decay, and the AdamW optimizer~\citep{Loshchilov2017FixingWD}. The effective batch size and peak learning rate are 128 and 1e-4, respectively. Validation occurs twice per epoch, and training stops after 10 consecutive validation checks without performance improvement. For more details, refer to AutoMM’s codebase\footnote{https://github.com/autogluon/autogluon/tree/master/multimodal/src/autogluon/multimodal}. Note that this naive baseline does not include some training procedure refinements like learning rate warmup, as they serve as the basic tricks in our examinations.

% The model checkpoint with the best validation performance is selected as the final model.
% \vspace{-1em}
\subsection{Basic Tricks}
On top of the vanilla baseline, we examine several heuristics for refining the training procedure. The heuristics we evaluate include:

\begin{itemize}
\setlength\itemsep{0em}
    \item Greedy soup~\citep{wortsman2022model}: Averaging the weights of multiple fine-tuned models, known as model soup, often improves accuracy and robustness. The greedy soup method sequentially adds models to the soup, retaining only those that enhance performance on a validation set, and outperforms uniform averaging. Models are sorted in decreasing order of validation set accuracy before running this procedure. This paper applies greedy soup to the top 3 checkpoints along a single training trajectory.
    \item Gradient clipping~\citep{zhang2019gradient}: This technique addresses the exploding gradient problem by limiting the magnitude of the gradients, preventing them from growing unchecked during training. In our experiments, we clip the gradients with a norm threshold of 1.0.
    \item Learning rate warmup~\citep{Liu2019OnTV}: Starting with a small learning rate and gradually increasing it over a few initial epochs or iterations helps prevent the model from diverging during the initial phase of training. We use linear warmup, which linearly increases the learning rate from 0 to the peak learning rate during the first 10\% of training steps.
    
    % This method is often used in training deep networks from scratch or fine-tuning pre-trained models. 
    % \item Cosine learning rate decay: When training deep networks, it is beneficial to anneal the learning rate over time. The intuition is that a high learning rate can introduce too much kinetic energy into optimization, making it difficult to converge to an optimal point. Immediately after the warmup stage, we steadily decrease the learning rate from the peak value to 0 following the cosine function.
    % \item Weight decay without bias and normalization: Weight decay is a popular regularization technique used in deep networks to improve generalization. In its original form, the model weights decay exponentially over the training steps. While equivalent to standard L2 regularization for stochastic gradient descent, this equivalence does not hold for adaptive gradient methods such as Adam. Instead of implementing it as the L2 regularization, AdamW decouples weight decay from gradient computation, preserving the original weight decay formula. Additionally, empirical evidence suggests not applying weight decay to bias parameters and normalization layers. In our experiments, we use a decay strength of 1e-3.
    \item Layerwise learning rate decay: It allocates distinct learning rates to each layer, with shallower layers assigned higher learning rates and smaller rates for deeper layers. This strategy is inspired by the premise that early layers learn general feature representations from large-scale pre-training, necessitating minor adjustments during fine-tuning. We use a decay rate of 0.9 in the experiments.

    % Later layers need more substantial adaptation to align with the specific downstream task, hence they are given larger learning rates.
    
\end{itemize}

\begin{table}[t]\Huge
    \caption{Results of adding the basic tricks incrementally. ``Baseline'' denotes that none of the tricks are applied. ``avg'' means the average score across datasets, and ``mrr'' shows the mean reciprocal rank among all evaluated methods, which indicates the frequency that one method outperforms others.}
    % \vspace{0.2em}
    \centering
    \resizebox{\textwidth}{!}{%
    \renewcommand\arraystretch{1.3}
    \begin{tabular}{lcccccccccc}
        \toprule
         \multirow{2}*{\bf Basic Trick} & \multicolumn{2}{c}{\bf Text+Tabular} & \multicolumn{2}{c}{\bf Image+Text} & \multicolumn{2}{c}{\bf Image+Tabular} & \multicolumn{2}{c}{\bf Image+Text+Tabular} & \multirow{2}*{\bf avg $\uparrow$} & \multirow{2}*{\bf mrr $\uparrow$}\\
         \cmidrule(lr){2-3} \cmidrule(lr){4-5}  \cmidrule(lr){6-7}  \cmidrule(lr){8-9}  
         % \cline{2-3} \cline{4-5}
         & \textbf{avg} & \textbf{mrr} & \textbf{avg } & \textbf{mrr}  & \textbf{avg} & \textbf{mrr}  & \textbf{avg} & \textbf{mrr}   \\
 
         % & roc\_auc & acc & acc & r2 & r2 & r2 &  \\
        \midrule
        Baseline & 0.525$_{0.039}$ & 0.376$_{0.017}$  & 0.655$_{0.007}$ & 0.420$_{0.038}$ & 0.805$_{0.000}$ & 0.272$_{0.032}$ & 0.515$_{0.011}$ & 0.280$_{0.058}$  & 0.615$_{0.014}$ & 0.336$_{0.014}$ \\
        
        +Greedy Soup  & 0.545$_{0.039}$ & 0.381$_{0.028}$  & 0.670$_{0.019}$ & 0.383$_{0.036}$  & 0.807$_{0.002}$ & 0.308$_{0.007}$  & 0.526$_{0.009}$ & 0.320$_{0.051}$  & 0.628$_{0.015}$ & 0.348$_{0.005}$ \\
 
        +Gradient Clip & 0.594$_{0.002}$ & 0.492$_{0.136}$  & 0.688$_{0.021}$ & 0.669$_{0.143}$  & 0.821$_{0.002}$ & 0.354$_{0.009}$  & 0.562$_{0.005}$ & 0.411$_{0.034}$  & 0.658$_{0.004}$ & 0.479$_{0.034}$ \\

        +LR Warmup  & 0.595$_{0.002}$ & 0.469$_{0.106}$  & \textbf{0.705$_{0.006}$} & \textbf{0.727$_{0.233}$}  & 0.829$_{0.001}$ & \textbf{0.763$_{0.107}$}  & 0.588$_{0.005}$ & 0.645$_{0.017}$  & 0.671$_{0.001}$ & \textbf{0.642$_{0.028}$} \\

        +Layerwise Decay  & \textbf{0.600$_{0.003}$} & \textbf{0.599$_{0.145}$}  & 0.701$_{0.003}$ & 0.570$_{0.099}$  & \textbf{0.830$_{0.004}$} & 0.717$_{0.062}$ & \textbf{0.590$_{0.003}$} & \textbf{0.648$_{0.052}$}  & \textbf{0.672$_{0.002}$} & 0.633$_{0.089}$ \\

        \bottomrule
    \end{tabular}
    }

    \label{tab:basic_tricks}
    \vspace{-0.5em}
\end{table}

In \Cref{tab:basic_tricks}, we find that the model average performance improves significantly when employing greedy soup, gradient clipping and learning rate warmup. The layerwise decay has mixed, yet generally positive,  effects when considering all the modality combinations. We refer to the baseline combined with all the basic tricks as \textbf{``Baseline+''}, which will be used in the following sections.

% When all the tricks are applied, both the average performance and the frequency of outperforming others are the highest.

\subsection{Multimodal Fusion Strategies}
\label{sec:struc}

% How to effectively combining different modalities is a key design challenge. The fusion method must be scalable to handle arbitrary combinations of images, text, and tabular data.

Multimodal classification and regression require fusing information from heterogeneous modalities to make predictions. Fusion approaches~\citep{vielzeuf2018centralnet,gadzicki2020early,zhang2021deep,boulahia2021early,yao2022modality,zhang2023meta,swamy2024multimodn}can be broadly categorized as early or late, depending on where the fusion occurs within the pipeline. Since we focus on fine-tuning pre-trained models~\citep{Dosovitskiy2020AnII,Vaswani2017AttentionIA}, we classify a strategy as early-fusion if the fusion occurs within a pre-trained model. Otherwise, it is considered late-fusion, as an extra fusion module is required on top of the pre-trained models. Here are the fusion alternatives we evaluate:

% , followed by a linear prediction head. All the pre-trained models are fine-tuned during training.

% Recently, Multimodal Large Language Models (MLLMs) have emerged as a research hotspot, leveraging powerful LLMs to perform multimodal tasks.
\begin{itemize}
\setlength\itemsep{0em}
    \item Late-fusion (LF): This approach uses unimodal (pre-trained) backbones to encode each modality independently. The encoded features are concatenated and merged using a fusion module. We investigate four variants of late fusion:
    \begin{itemize}
        \item LF-MLP: Uses Swin-Transformer Large, DeBERTaV3, and FT-Transformer to encode the image, text, and tabular data, respectively. An MLP module is used for fusing the concatenated CLS~\citep{Devlin2019BERTPO} token features.
        \item LF-Transformer: Uses the same unimodal encoders as LF-MLP but replaces the MLP with a transformer-based fusion module~\citep{Vaswani2017AttentionIA}.
        \item LF-Aligned: Follows the LF-MLP setup but leverages CLIP’s image and text encoders, which have been aligned through large-scale contrastive pre-training~\citep{radford2021learning}. We use the CLIP ViT-L/14 variant in the experiments.
        \item LF-LLM: Also follows LF-MLP but changes the fusion module to a Large Language Model (LLM)~\citep{Brown2020LanguageMA}. A typical Multimodal Large Language Models (MLLMs)~\citep{Yin2023ASO} architecture includes unimodal encoders, a connector, and an LLM. While MLLMs are primarily used for generative tasks, we are also interested in their performance on discriminative tasks. Due to the structural similarities between MLLMs and late fusion, we use the LLM as the fusion module, specifically employing the 7B version of Llama2~\citep{touvron2023llama} with the parameter-efficient tuning method QLoRA~\citep{dettmers2024qlora}.
        \item LF-SF (Sequential Fusion): Instead of fusing all the features simultaneously, sequential fusion~\citep{swamy2024multimodn} encodes and merges features from different modalities one after the other, similar to a Recurrent Neural Network (RNN).
    \end{itemize}

    % In addition to the parallel fusion approaches mentioned above, sequential fusion is another type of late fusion. 
    % For instance, MultiModN uses a state variable to represent the fused features, which is then passed to subsequent encoders for further fusion.
    
\item Early-fusion: A fusion module within the pre-trained model learns the joint representation of all modalities. The modalities can either be concatenated at the input level and passed through a shared encoder or processed separately and then fused later within the pre-trained model. We use Meta-Transformer~\citep{zhang2023meta}, the first framework to perform unified learning with data across 12 modalities, including images, text, and tabular data. In our experiments, we fine-tune its pre-trained modality-shared encoder.
\end{itemize}

\begin{table}[t]\Huge
    \caption{Results of multimodal fusion strategies. }
    % \vspace{0.2em}
    \centering
    \resizebox{\textwidth}{!}{%
    \renewcommand\arraystretch{1.5}
    \begin{tabular}{lcccccccccc}
        \toprule
         \multirow{2}*{\bf Fusion Struc.} &  \multicolumn{2}{c}{\bf Text+Tabular} & \multicolumn{2}{c}{\bf Image+Text} & \multicolumn{2}{c}{\bf Image+Tabular} & \multicolumn{2}{c}{\bf Image+Text+Tabular} & \multirow{2}*{\bf avg $\uparrow$} & \multirow{2}*{\bf mrr $\uparrow$}\\
         \cmidrule(lr){2-3} \cmidrule(lr){4-5}  \cmidrule(lr){6-7}  \cmidrule(lr){8-9} 
         & \textbf{avg} & \textbf{mrr} & \textbf{avg} & \textbf{mrr} & \textbf{avg} & \textbf{mrr} & \textbf{avg} & \textbf{mrr} \\

         % & roc\_auc & acc & acc & r2 & r2 & r2 &  \\
        \midrule
         LF-MLP(Baseline+) &  0.600$_{0.003}$ & 0.597$_{0.049}$  & 0.701$_{0.003}$ & 0.399$_{0.043}$  & 0.830$_{0.004}$ & 0.374$_{0.069}$ & 0.590$_{0.003}$ & 0.546$_{0.047}$  & 0.672$_{0.002}$ & 0.488$_{0.044}$ \\
         
         LF-Transformer & \textbf{0.601$_{0.001}$} & \textbf{0.644$_{0.062}$}  & 0.704$_{0.007}$ & 0.354$_{0.063}$  & 0.828$_{0.002}$ & 0.392$_{0.083}$ & 0.560$_{0.012}$ & 0.331$_{0.037}$ & 0.665$_{0.002}$ & 0.435$_{0.033}$ \\
         
         LF-Aligned & 0.538$_{0.005}$ & 0.240$_{0.027}$ & \textbf{0.774$_{0.002}$} & \textbf{0.891$_{0.077}$}  & \textbf{0.848$_{0.004}$} & \textbf{0.839$_{0.064}$}  & \textbf{0.616$_{0.008}$} & \textbf{0.870$_{0.007}$}  & \textbf{0.683$_{0.001}$} & \textbf{0.696$_{0.033}$} \\
         
         LF-LLM & 0.463$_{0.003}$ & 0.261$_{0.059}$  & 0.698$_{0.009}$ & 0.389$_{0.057}$  & 0.826$_{0.002}$ & 0.411$_{0.016}$   & 0.556$_{0.008}$ & 0.361$_{0.020}$  & 0.625$_{0.004}$ & 0.352$_{0.019}$ \\
         
         LF-SF & 0.585$_{0.008}$ & 0.544$_{0.014}$  & 0.427$_{0.072}$ & 0.322$_{0.042}$ & 0.665$_{0.007}$ & 0.208$_{0.024}$  & 0.295$_{0.03}$ & 0.174$_{0.003}$  & 0.488$_{0.016}$ & 0.316$_{0.001}$ \\
         
         Early-fusion &  0.484$_{0.012}$ & 0.195$_{0.012}$ & 0.579$_{0.002}$ & 0.334$_{0.057}$ & 0.747$_{0.057}$ & 0.273$_{0.118}$  & 0.444$_{0.007}$ & 0.208$_{0.007}$  & 0.555$_{0.015}$ & 0.248$_{0.040}$\\
         
        \bottomrule
    \end{tabular}
    }
    
    \label{tab:struc}
    \vspace{-0.5em}
\end{table}

According to \Cref{tab:struc}, late-fusion strategies, except LF-SF, generally outperform early-fusion across all datasets. Improving early-fusion pre-training for such multimodal downstream tasks remains an open problem. Among the late fusion methods, LF-Aligned significantly outperforms other variants across different modality compositions, except for the text+tabular composition. This underscores the importance of cross-modal alignment in pre-training and implies that aligning tabular data with other modalities in pre-training may yield further benefits.

LF-MLP and LF-Transformer show comparable performance on most compositions, while LF-MLP significantly outperforms LF-Transformer in the image+text+tabular composition. This suggests that an MLP might be a better choice than a transformer as the fusion module when the number of modalities increases.  Additionally, the performance of LF-LLM and LF-SF is subpar. We attribute the failure of LF-LLM to its pre-training for generative tasks, which may not adapt well to downstream discriminative tasks without higher training costs, including more trainable parameters and additional fine-tuning data. And LF-SF uses a state variable to represent the multimodal fusion features, passing it to the next encoders, which may result in the loss of information from earlier modalities. 

% This suggests that concatenating features along the channel dimension is more effective than along the token dimension in this case. Explicit fusion across the channels of multimodal features appears to be more beneficial as the number of modalities increases. 

% Additionally, the performance of LF-LLM and LF-SF is subpar. We attribute the failure of LF-LLM to the fact that the LLM module LLAMA2 used by LF-LLM is primarily designed for generative tasks. Adapting it to downstream discriminative tasks may require higher training costs, including more trainable parameters and additional fine-tuning data. And LF-SF uses a state variable to represent the multimodal fusion feature, passing it to the next encoder. This may result in the loss of information from earlier modalities. 
% training time?

\subsection{Converting Tabular Data into Text}

Tabular data remains widely used across various ML tasks in both science and industry. Recently, owing to the remarkable emergent capabilities of LLMs~\citep{Yin2023ASO}, several studies ~\citep{carballo2022tabtext,hegselmann2023tabllm,nam2023semi,wang2023unipredict} have applied LLMs to tabular data classification and cleaning. Utilizing an LLM for tabular data necessitates serializing the data into a natural text representation. However, tabular serialization—converting tabular data into text—in multimodal settings remains relatively under-explored. In our experiments, we serialize categorical and numeric values separately to assess their individual effects. We explore several serialization methods~\citep{jaitly2023towards}, but due to space constraints, we present only the best serialization results here. For a comprehensive comparison of all serialization methods, refer to \Cref{tab:serial_methods}.

\begin{table}[t]
    \caption{Results of converting tabular data into text data on \textit{datasets with tabular data}.}
    % \vspace{0.2em}
    \centering
    \resizebox{\textwidth}{!}{%
    \renewcommand\arraystretch{1.3}
    \begin{tabular}{lcccccccc}
        \toprule
         \multirow{2}*{\bf Method.} &  \multicolumn{2}{c}{\bf Text+Tabular} &  \multicolumn{2}{c}{\bf Image+Tabular} & \multicolumn{2}{c}{\bf Image+Text+Tabular} & \multirow{2}*{\bf avg $\uparrow$} & \multirow{2}*{\bf mrr $\uparrow$}\\
         \cmidrule(lr){2-3} \cmidrule(lr){4-5}  \cmidrule(lr){6-7}   
         &  \textbf{avg} & \textbf{mrr} & \textbf{avg} & \textbf{mrr} &  \textbf{avg} & \textbf{mrr} \\

         % & roc\_auc & acc & acc & r2 & r2 & r2 &  \\
        \midrule
         Baseline+ & 0.600$_{0.003}$ & 0.713$_{0.151}$  & 0.830$_{0.004}$ & 0.689$_{0.177}$ & \textbf{0.590$_{0.003}$} & \textbf{0.713$_{0.069}$}  & 0.664$_{0.002}$ & \textbf{0.706$_{0.097}$} \\
       
        Convert Categorical  & \textbf{0.602$_{0.001}$} & \textbf{0.731$_{0.073}$}  & \textbf{0.837$_{0.001}$} & \textbf{0.800$_{0.216}$}  & 0.585$_{0.006}$ & 0.602$_{0.052}$ & \textbf{0.665$_{0.002}$} & \textbf{0.706$_{0.081}$} \\
        
         Convert Numeric & 0.597$_{0.001}$ & 0.611$_{0.068}$  & 0.806$_{0.003}$ & 0.522$_{0.103}$ & 0.579$_{0.004}$ & 0.565$_{0.026}$ & 0.652$_{0.001}$ & 0.569$_{0.037}$ \\
         
        \bottomrule
    \end{tabular}
    }

    \label{tab:covert_tabular}
\end{table}
In \Cref{tab:covert_tabular}, converting numeric values into text leads to a negative performance effect compared to Baseline+. This may be because converting precise numeric values into text strings makes it more challenging for the model to learn inherent comparative relationships. However, converting categorical values can enhance model performance in most cases, except for the image+text+tabular composition. We attribute this improved performance to the similarity between categorical and text data. Representing categorical data in text format can directly convey its semantic information, potentially helping the model better understand the meaning of different categories in most cases.

% In \Cref{tab:covert_tabular}, Convert Numerical leads to a negative performance effect compared to using the original numerical values across all benchmarks. Numerical data reflects the comparative relationships between samples, which is helpful for model decision making. For instance, a model can directly learn the comparative relationships among different pets' weights based on numerical values. However, converting precise numerical values into text strings makes it more challenging for the model to learn the inherent comparative relationships than using the numerical values directly. 

% Converting Categorical does not improve model performance in the Image+Text+Tabular composition but does enhance performance in Text+Tabular and Image+Tabular compositions. We attribute the improved model performance to the similarity between categorical and text data. Representing categorical data in text format can directly convey its semantic information, potentially helping the model better understand the meaning of different categories in most cases. 

\subsection{Cross-modal Alignment}

% Aligning different modalities through contrastive cross-modal pre-training has achieved remarkable success in tasks such as cross-modal retrieval and zero-shot classification.

The goal of cross-modal alignment~\citep{radford2021learning} is to project data from different modalities into a shared embedding space, where paired samples are brought closer, and unpaired samples are pushed apart.  While \Cref{sec:struc} demonstrates that aligned pre-trained unimodal encoders perform better than independent ones during fine-tuning, it is unclear whether incorporating an extra alignment objective during supervised fine-tuning is also beneficial. We explore two alignment objectives: one considering positive-only pairs, and the other considering both positive and negative pairs. The positive-only approach minimizes the KL divergence~\citep{Csiszr1975IDivergenceGO} losses among different modalities within each sample. In contrast, the positive+negative approach uses the InfoNCE~\citep{Oord2018RepresentationLW} loss, which additionally separates modalities across different samples.

\begin{table}[t]\Huge
    \caption{Results of cross-modal alignment. }
    % \vspace{0.2em}
    \centering
    \resizebox{\textwidth}{!}{%
    \renewcommand\arraystretch{1.5}
    \begin{tabular}{lcccccccccc}
        \toprule
         \multirow{2}*{\bf Method}  & \multicolumn{2}{c}{\bf Text+Tabular} & \multicolumn{2}{c}{\bf Image+Text} & \multicolumn{2}{c}{\bf Image+Tabular} & \multicolumn{2}{c}{\bf Image+Text+Tabular} & \multirow{2}*{\bf avg $\uparrow$} & \multirow{2}*{\bf mrr $\uparrow$}\\

        \cmidrule(lr){2-3} \cmidrule(lr){4-5}  \cmidrule(lr){6-7}  \cmidrule(lr){8-9}

         & \textbf{avg} & \textbf{mrr} & \textbf{avg} & \textbf{mrr} & \textbf{avg} & \textbf{mrr} & \textbf{avg} & \textbf{mrr} &\\
 
         % & roc\_auc & acc & acc & r2 & r2 & r2 &  \\
        \midrule
         Baseline+ & \textbf{0.600$_{0.003}$} & 0.648$_{0.170}$  & 0.701$_{0.003}$ & 0.422$_{0.079}$  & 0.830$_{0.004}$ & 0.611$_{0.140}$  & 0.590$_{0.003}$ & 0.556$_{0.023}$  & 0.672$_{0.002}$ & 0.563$_{0.090}$ \\
         
         Positive-only & 0.599$_{0.002}$ & \textbf{0.657$_{0.035}$}  & 0.707$_{0.007}$ & 0.711$_{0.042}$ & \textbf{0.831$_{0.002}$} & \textbf{0.778$_{0.096}$}  & \textbf{0.602$_{0.007}$} & \textbf{0.824$_{0.013}$} & \textbf{0.677$_{0.002}$} & \textbf{0.742$_{0.031}$} \\
         
         Positive+Negative  & \textbf{0.600$_{0.002}$} & 0.611$_{0.138}$  & \textbf{0.720$_{0.007}$} & \textbf{0.856$_{0.042}$} & 0.823$_{0.002}$ & 0.478$_{0.031}$  & 0.581$_{0.002}$ & 0.481$_{0.057}$ & 0.673$_{0.002}$ & 0.601$_{0.029}$ \\
       
        \bottomrule
    \end{tabular}
    }

    \label{tab:align}
\end{table}

In \Cref{tab:align}, incorporating an additional alignment objective during fine-tuning enhances the model's performance on average. Specifically, the positive-only approach demonstrates comparable or improved performance compared to Baseline+ across all modality compositions. However, while the positive+negative approach shows significant performance improvement on the image+text combination, it still lags behind Baseline+ on image+tabular and image+text+tabular compositions. This may be due to the lack of enough negative samples, as the batch size is kept at 128 in our experiments. Sufficient negative samples are crucial for the model to effectively discriminate between positive and negative samples when using contrastive learning ~\citep{he2020momentum,chen2020simple,chen2020improved}.

% In \Cref{tab:align}, incorporating an additional alignment objective during fine-tuning enhances the model's performance on average. This further demonstrates the effectiveness of cross-modal alignment, which is beneficial not only during the model's initialization but also during fine-tuning. Specifically, the Positive-only approach demonstrates comparable or improved performance compared to Baseline+ across all the modality compositions. However, while the Positive+Negative approach shows significant performance improvement on Image+Text compositions, it still lags behind Baseline+ on Image+Tabular and Image+Text+Tabular compositions. We attribute this to the need for more negative samples for these two modality compositions. The model may require more negative samples to learn better features; otherwise, performance could be negatively impacted (the batch size is kept at 128 in our case). Sufficient negative samples are crucial for the model to effectively discriminate between positive and negative samples when using this approach~\citep{he2020momentum,chen2020simple,chen2020improved}.

\subsection{Multimodal Data Augmentation}

% It allows models to see more diverse features and improve generalization. 

Data augmentation enriches datasets by creating variations of existing data, serving as a crucial training technique in modern deep learning. While data augmentation is well-studied for unimodal data such as images~\citep{Cubuk2019AutoAugmentLA, Lim2019FastA} and text~\citep{wei2019eda, Karimi2021AEDAAE}, it has received less attention in the context of multimodal data. Using our multimodal supervised learning benchmark, we compare several multimodal augmentation methods, categorized as follows:
\begin{itemize}
\setlength\itemsep{0em}
    \item Input Augmentation (Input Aug.): It applies augmentation operations directly to the input data. Each modality in a sample is independently augmented on-the-fly during training. Specifically, we use TrivialAugment~\citep{muller2021trivialaugment} for image data and EDA~\citep{wei2019eda} for text data. Due to the lack of suitable transformations, tabular data remain unaltered.
    \item Feature Augmentation: It augments samples in the feature space within the model. Augmentation can be independent or joint with respect to the different modalities.
    \begin{itemize}
        \item Independent Feature Augmentation (Feature Aug.(Inde.)): A representative method is Manifold Mixup~\citep{verma2019manifold}, which interpolates latent vectors and labels to generate new samples. We apply Manifold Mixup to each modality independently.
        \item Joint Feature Augmentation (Feature Aug.(Joint)): A state-of-the-art approach is LeMDA~\citep{liu2022learning}, which trains an augmentation network to learn how to jointly augment all the modalities of each sample in the feature space.
    \end{itemize}
\end{itemize}

\begin{table}[t]\Huge
    \caption{Results of multimodal data augmentation.}
    \centering
    \resizebox{\textwidth}{!}{%
    \renewcommand\arraystretch{1.3}
    \begin{tabular}{lcccccccccc}
        \toprule
         \multirow{2}*{\bf Method} &\multicolumn{2}{c}{\bf Text+Tabular} & \multicolumn{2}{c}{\bf Image+Text} & \multicolumn{2}{c}{\bf Image+Tabular} & \multicolumn{2}{c}{\bf Image+Text+Tabular} & \multirow{2}*{\bf avg$\uparrow$} & \multirow{2}*{\bf mrr$\uparrow$}\\

        \cmidrule(lr){2-3} \cmidrule(lr){4-5}  \cmidrule(lr){6-7}  \cmidrule(lr){8-9} 

         & \textbf{avg} & \textbf{mrr} & \textbf{avg} & \textbf{mrr} & \textbf{avg} & \textbf{mrr} & \textbf{avg} & \textbf{mrr}  \\
 
         % & roc\_auc & acc & acc & r2 & r2 & r2 &  \\
        \midrule
         Baseline+ & 0.600$_{0.003}$ & 0.495$_{0.195}$  & 0.701$_{0.003}$ & 0.417$_{0.131}$  & 0.830$_{0.004}$ & 0.572$_{0.157}$  & 0.590$_{0.003}$ & 0.593$_{0.094}$  & 0.672$_{0.002}$ & 0.521$_{0.116}$\\
         
         Input Aug.& 0.601$_{0.003}$ & 0.426$_{0.064}$  & \textbf{0.710$_{0.004}$} & \textbf{0.683$_{0.049}$} & \textbf{0.832$_{0.002}$} & \textbf{0.800$_{0.054}$}  & \textbf{0.594$_{0.002}$} & \textbf{0.731$_{0.076}$} & \textbf{0.676$_{0.001}$} & \textbf{0.653$_{0.018}$} \\
         
         Feature Aug.(Inde.) & 0.601$_{0.002}$ & 0.426$_{0.080}$  & 0.706$_{0.006}$ & 0.594$_{0.102}$ & 0.805$_{0.003}$ & 0.333$_{0.059}$ & 0.550$_{0.006}$ & 0.329$_{0.043}$  & 0.657$_{0.001}$ & 0.417$_{0.047}$ \\
         
         Feature Aug.(Joint)  & \textbf{0.604$_{0.003}$} & \textbf{0.741$_{0.125}$} & 0.705$_{0.010}$ & 0.456$_{0.245}$  & 0.826$_{0.002}$ & 0.456$_{0.087}$ & 0.586$_{0.005}$ & 0.481$_{0.114}$  & 0.672$_{0.001}$ & 0.540$_{0.056}$ \\
        
        \bottomrule
    \end{tabular}
    }
    \label{tab:aug}
\end{table}

In \Cref{tab:aug}, Input Aug. performs best in most cases. Both Feature Aug.(Inde.) and Feature Aug.(Joint) demonstrate mixed performance, but Feature Aug.(Joint) achieves the best performance on the text+tabular datasets.

% In \Cref{tab:aug}, Input Aug. performs best in Image+Text composition, and improves average performance compared to not applying data augmentation. There are few suitable methods for augmenting tabular data, but if an effective tabular input augmentation method were used, the performance may be further enhanced. Feature Aug.(Inde.) and Feature Aug.(Joint) improve average performance compared to no-augmentation baseline on the Text+Tabular and Image+Tabular compostions. However, they do not outperform the no-augmentation baseline on Image+Tabular and Image+Text+Tabular compositions. 

\subsection{Handling Modality Missingness}

\begin{table}[t]\Huge
    \caption{Results of handling modality missingness on \textit{partial datasets with missingness}. }
    \centering
    \resizebox{\textwidth}{!}{%
    \renewcommand\arraystretch{1.5}
    \begin{tabular}{lcccccccccc}
        \toprule
         \multirow{2}*{\bf Method.} &\multicolumn{2}{c}{\bf Text+Tabular} & \multicolumn{2}{c}{\bf Image+Text} & \multicolumn{2}{c}{\bf Image+Tabular} & \multicolumn{2}{c}{\bf Image+Text+Tabular} & \multirow{2}*{\bf avg$\uparrow$} & \multirow{2}*{\bf mrr$\uparrow$}\\

           \cmidrule(lr){2-3} \cmidrule(lr){4-5}  \cmidrule(lr){6-7}  \cmidrule(lr){8-9}

         & \textbf{avg} & \textbf{mrr}  & \textbf{avg} & \textbf{mrr} & \textbf{avg} & \textbf{mrr} & \textbf{avg} & \textbf{mrr}  \\
 
         % & roc\_auc & acc & acc & r2 & r2 & r2 &  \\
        \midrule
         Baseline+ & 0.708$_{0.003}$ & 0.546$_{0.182}$ & 0.732$_{0.003}$ & 0.583$_{0.297}$ & 0.825$_{0.004}$ & 0.593$_{0.069}$ & 0.673$_{0.001}$ & 0.514$_{0.103}$ & \textbf{0.730$_{0.002}$} & 0.553$_{0.083}$ \\

         Modality Dropout  & 0.714$_{0.003}$ & 0.833$_{0.236}$ & 0.732$_{0.001}$ & 0.539$_{0.086}$  & 0.812$_{0.004}$ & 0.359$_{0.128}$ & 0.678$_{0.006}$ & 0.515$_{0.087}$  & \textbf{0.730$_{0.002}$} & 0.560$_{0.059}$ \\
         
         Learnable Embed(Numeric)& 0.708$_{0.005}$ & 0.467$_{0.126}$ & 0.732$_{0.003}$ & 0.583$_{0.297}$  & 0.824$_{0.002}$ & 0.531$_{0.137}$  & 0.671$_{0.002}$ & 0.436$_{0.059}$  & 0.728$_{0.002}$ & 0.492$_{0.052}$ \\
         
         Learnable Embed(Image) & 0.708$_{0.003}$ & 0.546$_{0.182}$ & 0.731$_{0.001}$ & 0.569$_{0.109}$  & \textbf{0.827$_{0.008}$} & \textbf{0.800$_{0.178}$}  & 0.674$_{0.004}$ & 0.517$_{0.080}$  & \textbf{0.730$_{0.003}$} & \textbf{0.604$_{0.060}$} \\
         
         Modality Drop.+Learn. Embed(Image)  & \textbf{0.714$_{0.003}$} & \textbf{0.833$_{0.236}$}  & \textbf{0.736$_{0.002}$} & \textbf{0.681$_{0.255}$} & 0.807$_{0.012}$ & 0.265$_{0.068}$  & \textbf{0.679$_{0.006}$} & \textbf{0.603$_{0.030}$} & 0.729$_{0.002}$ & 0.589$_{0.070}$ \\

        \bottomrule
    \end{tabular}
    }

    \label{tab:missingness}
\end{table}

In real-world applications, it is common for some samples to have incomplete modalities during either training or deployment~\citep{ma2021smil,zhao2021missing,zeng2022tag,lee2023multimodal}. Therefore, it is crucial to equip models with the ability to handle missing modalities. We find that AutoMM~\citep{tang2024autogluon}, by default, can address modality missingness. Specifically, it fills missing numeric fields in a column with the column's mean value, assigns a new category for missing categorical fields in one column, uses a zero image for missing images, and represents an empty text sequence with the CLS~\citep{Devlin2019BERTPO} token. We further examine two additional techniques for handling modality missingness, applicable to image, text, and tabular data.
% To make the model robust to missing modalities, it can be beneficial to expose it to many samples with missing modalities during training. Modality dropout, i
\begin{itemize}
\setlength\itemsep{0em}
    \item Modality Dropout: Inspired by Dropout~\citep{srivastava2014dropout}, it randomly discards each modality of each training sample with a predefined probability.
    \item Learnable Embeddings~\citep{lee2023multimodal}: AutoMM's default setup already employs learnable embeddings for missing categorical and text fields, as the new category has corresponding learnable embeddings and the CLS token’s embeddings are fine-tuned during training. Building on this setup, we explore learnable embeddings for missing numeric and image fields. For better comparison with the baseline, we initialize their learnable embeddings using the mean numeric value and a zero image, respectively.
\end{itemize}

We conduct experiments on partial datasets with missing modalities; details are provided in \Cref{tab:missing_dataset_info}. In \Cref{tab:missingness}, modality dropout improves model performance on the text+tabular and image+text+tabular datasets. While learnable image embedding enhances performance in image+tabular and image+text+tabular compositions, learnable numeric embedding shows no improvement. Moreover, the joint use of modality dropout and learnable image embedding achieves the best results in most cases but falls behind learnable image embedding alone on image+tabular datasets, indicating that simply stacking these two techniques is not universally effective.

% We conduct experiments on partial datasets with modality missingness, refer to the appendix for the details of these datasets. In \Cref{tab:missingness}, Modality Dropout improves the performance compared to Baseline+ in Text+Tabular and Image+Text compositions. And Learnable Embedding for image data improves performance in Image+Tabular and Image+Text+Tabular compositions, while Learnable Embedding for numerical data shows no improvement in any composition. We further combine Modality Dropout and Learnable Embedding for image data to evaluate their performance. Modality Drop.+Learn. Embed(Image) achieves the best results in Text+Tabular, Image+Text, and Image+Text+Tabular compositions. However, it performs the worst in the Image+Tabular composition, indicating that simply stacking these two techniques is not universally effective.

% Modality dropout improves performance on datasets both with and without missing data. For datasets without missing data, this improvement is attributed to the regularization effect of modality dropout, which enhances model robustness and alleviates overfitting.

% To investigate a more generalized scenario where the missing ratios in the training and test sets differ, we randomly drop modalities at varying ratios in datasets without missing data.

% Modality dropout rate ablation:
% ...

% We divide the datasets to with or without missingness. For both datasets  we can see that. it is a regularization trick.

\subsection{Integrating Bag of Tricks}

We have investigated multiple techniques for enhancing multimodal supervised learning. This section details how to integrate these techniques into a single, robust modeling strategy that performs consistently across diverse datasets without manual adjustments. One approach is to stack all effective tricks into a single model (Stacking). For each modality combination, we incorporate all the tricks that enhance the Baseline+ model from previous sections. Additionally, the above experiments have created a model pool that includes Baseline+ and its variants with different advanced tricks. A simple approach is to average the predictions of all models in the pool (Average All). A more advanced method involves learning an ensemble~\citep{caruana2004ensemble}, assigning higher weights to important models, lower weights to less important models, and zero weights to ineffective models (Ensemble Selection).

\begin{figure}[t]
    \centering
    \includegraphics[width=\linewidth]{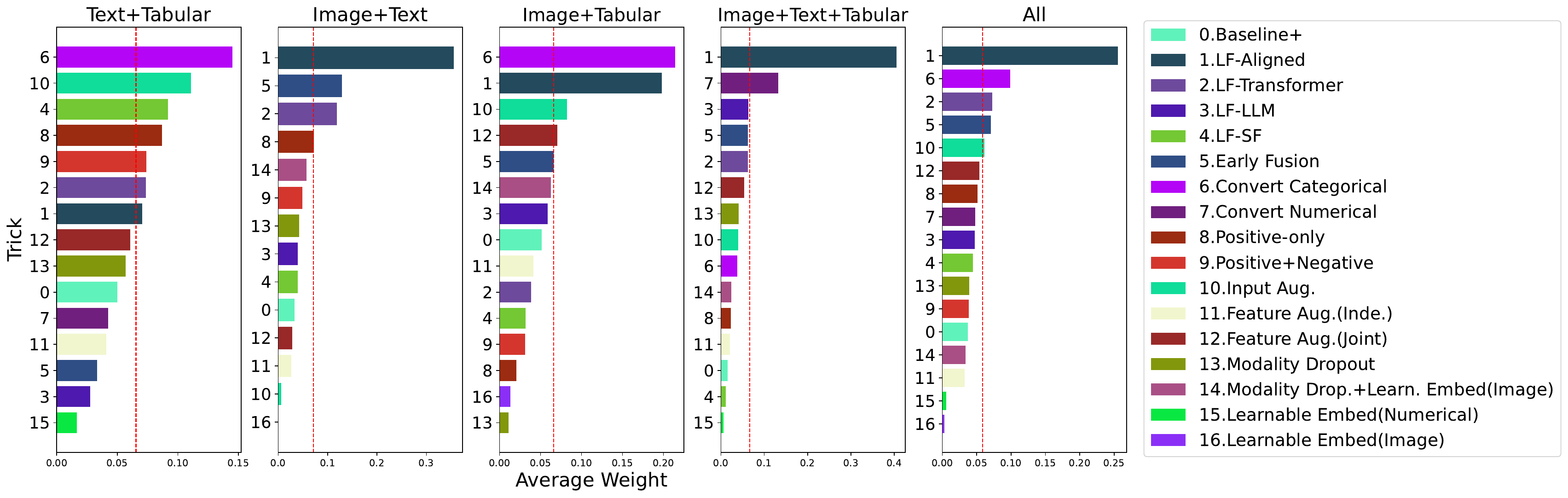}
    \caption{Weight statistics of tricks in Ensemble Selection. Each bar represents the average weight of a trick for a specific modality combination or across the entire benchmark (All). The red dashed lines indicate the average levels; bars above this line denote tricks with above-average importance.}
    % \vspace{-10pt}
    \label{fig:ensemble}
    \vspace{-0.5em}
\end{figure}

\begin{table}[t]\Huge
    \caption{Results of integrating bag of tricks. }
    \centering
    \resizebox{\textwidth}{!}{%
    \renewcommand\arraystretch{1.5}
    \begin{tabular}{lcccccccccc}
        \toprule
         \multirow{2}*{\bf Method}  & \multicolumn{2}{c}{\bf Text+Tabular} & \multicolumn{2}{c}{\bf Image+Text} & \multicolumn{2}{c}{\bf Image+Tabular} & \multicolumn{2}{c}{\bf Image+Text+Tabular} & \multirow{2}*{\bf avg $\uparrow$} & \multirow{2}*{\bf mrr $\uparrow$}\\

        \cmidrule(lr){2-3} \cmidrule(lr){4-5}  \cmidrule(lr){6-7}  \cmidrule(lr){8-9} 

         & \textbf{avg} & \textbf{mrr} & \textbf{avg} & \textbf{mrr} & \textbf{avg} & \textbf{mrr} & \textbf{avg} & \textbf{mrr} &\\
 
         % & roc\_auc & acc & acc & r2 & r2 & r2 &  \\
        \midrule
         Stacking & 0.599$_{0.004}$ & 0.519$_{0.052}$  & 0.770$_{0.004}$ & 0.589$_{0.057}$ & 0.865$_{0.002}$ & 0.600$_{0.072}$  & 0.609$_{0.005}$ & 0.426$_{0.035}$  & 0.701$_{0.002}$ & 0.528$_{0.023}$\\
         
         Average All &  0.619$_{0.000}$ & 0.667$_{0.039}$ & 0.741$_{0.004}$ & 0.578$_{0.031}$ & 0.854$_{0.001}$ & 0.489$_{0.016}$  & 0.625$_{0.002}$ & 0.491$_{0.035}$ & 0.702$_{0.001}$ & 0.558$_{0.022}$\\
         
         Ensemble Selection  & \textbf{0.620$_{0.001}$} & \textbf{0.676$_{0.086}$}  & \textbf{0.781$_{0.004}$} & \textbf{0.889$_{0.096}$}  & \textbf{0.865$_{0.004}$} & \textbf{0.800$_{0.082}$}  & \textbf{0.652$_{0.003}$} & \textbf{0.917$_{0.000}$}  & \textbf{0.721$_{0.001}$} & \textbf{0.818$_{0.021}$} \\

        \bottomrule
    \end{tabular}
    }
    
    \label{tab:integrate}

\end{table}

In \Cref{tab:integrate}, Ensemble Selection achieves stable and best performance across the entire benchmark, showcasing the effectiveness and necessity of dynamically weighting different models in ensembling. Unlike Average All, where all models contribute equally, Ensemble Selection can dynamically adjust model contributions, making it more adaptable to diverse real-world datasets. Compared to Stacking, Ensemble Selection eliminates the need to manually select models for each modality composition and mitigates potential conflicts between different tricks, which could otherwise harm performance.

% As shown in the figure, LF-Aligned, Convert Categorical, and LF-Tansformer are the main ensembling tricks when averaged across the total benchmark. For Text+Tabular composition, the main ensembling tricks are Convert Categorical, Input Aug., LF-SF, Positive-only, Positive+Negative, and LF-Transformer. For Image+Text composition, the main ensembling tricks are LF-Aligned, LF-Transformer and Positive+Negative. For Image+Tabular composition, the main ensembling tricks are LF-Aligned, Convert Categorical, and Feature Aug.(Joint). For Image+Text+Tabular composition, the main ensembling tricks are LF-Aligned, Convert Numerical and LF-Transformer.

We further investigate the contribution of each trick during Ensemble Selection to discern any underlying patterns. We identify the tricks that consistently contribute above-average levels during ensembling across the total benchmark and specific modality compositions. We call these ``main ensembling tricks". As shown in \Cref{fig:ensemble}, the main ensembling tricks are multimodal fusion strategies, converting tabular data into text, data augmentation, cross-modal alignment, and handling missing modalities, with their importance decreasing according to their average weights.

Specifically, most main ensembling tricks are those that yield positive improvements in our previous investigation, such as Convert Categorical and LF-Aligned. Interestingly, some tricks with negative impacts can also be main ensembling tricks, such as Convert Numeric in the image+text+tabular composition. This demonstrates ensembling's unique advantage over stacking: it can leverage even those tricks with negative impacts.

% Convert Categorical is the main ensembling trick in Text+Tabular and Image+Tabular modality compositions. For multimodal fusion strategies, LF-Aligned and LF-Transformer are the main ensembling tricks: LF-Aligned significantly contributes to Image+Text, Image+Tabular, and Image+Text+Tabular compositions, while LF-Transformer contributes to Text+Tabular, Image+Text, and Image+Text+Tabular compositions. For data augmentation, Positive-only is the main ensembling trick in Text+Tabular compostion, while Postive+Negative is the main ensembling trick in Text+Tabular and Image+Text compostions. For data augmentation, Input Aug. is the main ensembling trick in Text+Tabular composition, while Feature Aug.(Joint) is the main ensembling trick in Image+Tabular composition. 

% These tricks align with our expectations, as they show performance improvements compared to Baseline+ on modality-specific compositions in previous experiments. Therefore, if a limited number of models is given, we can prioritize training models equipped with these tricks in the order mentioned. 

% Surprisingly, Convert Numerical and LF-SF, which showed negative performance in earlier experiments, are main ensembling tricks in the Image+Text+Tabular and Text+Tabular compositions, respectively. This demonstrates ensembling's unique advantage over stacking: it can leverage even those tricks with negative impacts.

% \input{related_work_short}
\section{Conclusion and Future Work}

This paper examines various tricks to enhance multimodal supervised learning with flexible combinations of image, text, and tabular data. To this end, we curate a new benchmark with 22 datasets from diverse domains and real-world applications. Through ensemble selection, we effectively aggregate these diverse tricks into a unified pipeline, achieving optimal performance and uncovering several key insights:
\begin{itemize}
\setlength\itemsep{0em}
    \item The weights assigned to tricks suggest their relative importance: multimodal fusion strategy > converting tabular data to text > multimodal data augmentation > cross-modal alignment > handling missing modalities.
    \item Late fusion with aligned unimodal encoders and converting categorical data to text are the two most commonly employed tricks.
    \item While tricks that individually improve performance are more frequently selected for ensembles, some tricks with negative individual impacts, such as converting numeric data to text and late fusion with LLM, are also included at times, enhancing ensemble performance.
\end{itemize}

Moreover, the individual evaluation of each trick yields additional insights:
\begin{itemize}
\setlength\itemsep{0em}
    \item Basic tricks like greedy soup, gradient clipping, and learning rate warmup consistently improve performance.
    \item Late fusion using aligned unimodal encoders proves to be the most effective fusion strategy.
    \item Converting categorical data to text is beneficial, whereas converting numeric data is not.
    \item An additional cross-modal alignment objective boosts supervised fine-tuning performance.
    \item Despite its simplicity, input augmentation remains the superior data augmentation technique.
    \item Both modality dropout and learnable embeddings help handle the issue of missing modalities.
\end{itemize}

Despite our efforts to make the benchmarking comprehensive, some aspects of multimodal AutoML remain unexplored and are left for future work. These include balancing the importance of different modalities and selecting models from the numerous pre-trained options. Additionally, the data collected for text and tabular modalities is exclusively in English. Including low-resource languages would be valuable for creating a universal AutoML solution for multilingual applications. Extending the benchmark to encompass more modalities, such as audio and video, is another promising direction.

\bibliography{neurips_data_2024}
\bibliographystyle{plainnat}

\appendix
\newpage
\section{Limitations and Societal Impact}\label{sec:limitation_societal_impact}

Our benchmark only includes text in English, so its conclusions may be applicable only to English-language applications. To ensure similar advancements for text and tabular data in low-resource languages, we encourage the development of a comparable benchmark with non-English text. Additionally, multimodal memes from social media may contain hate speech. Detecting such content is crucial for mitigating the related societal issues. Furthermore, text fields can raise privacy concerns, as they may expose personal information. These fields can also introduce spurious correlations in training data, potentially harming accuracy during deployment and undesirably linking to protected attributes such as race, gender, or socioeconomic status.

\section{Benchmark Datasets Details}\label{sec:benchmark_details}

In developing a practical AutoML solution to real-world problems, the empirical performance of design decisions is what ultimately matters. The hope is that it can reliably produce reasonable accuracy on arbitrary datasets without manual user-tweaking. Since representative benchmarks comprised of many diverse datasets are critical for proper evaluation of AutoML, we introduce the first benchmark for evaluating multimodal image/text/tabular ML. Our benchmark completely contain all 4 combinations of image, text, and tabular data, with at 5 datasets for each modality combination, resulting totally 22 datasets. Note that all the datasets in our benchmark are public and available for research purpose.

% Our new benchmark is publicly available, as is the code to reproduce all results presented here.

Our benchmark spans a wide array of domains ranging from finance to healthcare, social media, food, transportation, arts, and e-commerce. The datasets reflect real-world applications in financial news classification, customer review score prediction, sentiment analysis of Internet memes, food category prediction, housing price prediction, skin lesion classification, artist category classification, pet adoption speed prediction, and traffic accident identification. Technically, the benchmark consists of a mix of classification and regression tasks and include a rich mix of image, text, categorical, and numeric columns. \Cref{tab:dataset_info} shows the datasets are quite diverse in terms of training sample number (1k - 20k), problem types (binary, multiclass, regression), number of features (0 - 47), and type of features (image+text, image+tabular, text+tabular, image+text+tabular). The text field length varies greatly from short product names to lengthy news report. 

To reflect real-world tasks, we preprocessed data minimally by ensuring the features/labels correspond to meaningful prediction tasks without duplicate examples. Thus, there are arbitrarily-formatted strings and missing values all throughout. Each dataset in our benchmark is provided with a pre-specified training/test split, which either uses the original split existed in a dataset or follows the 4:1 or 3:1 ratio. Following the principles of training ML models, the test data are not accessible during training. If the dataset does not contain validation data, we use a hold-out subset of training data as validation data for model selection and early stopping. Though the choice of training/validation split is a key design decision in AutoML, it is beyond the scope of this study. In our experiments, we always use the training/validation split provided by AutoMM, which is stratified based on labels in classification tasks.

% The benchmark GitHub repository contains: (i) methods to easily retrieve the individual datasets and train/test splits, (ii) code to run all of the tricks studied in this paper and reproduce our results, and (iii) the scripts we used to produce each benchmark dataset from the original data source. Common modifications made to original data sources to produce the benchmark dataset versions included: defining a practically meaningful prediction task if there was not one associated with the original dataset, omitting duplicated rows, omitting non-predictive features (e.g. user ID) and those that were too correlated with the prediction target (making the benchmark too easy otherwise), applying log-transform to prediction targets that correspond to product prices (log-scale errors are more meaningful in most real pricing applications), and down-sampling overly large datasets () to ensure the benchmark remains computationally accessible.

The details of the benchmark are as follows. For the Text+Tabular modality composition (fake, airbnb, channel, qaa, qaq, cloth), we directly follow the introductions and preprocessing steps in~\citep{shi2021benchmarking}.

\textbf{fake}: Predict whether online job postings are real or fake based on their text and additional tabular features like amount of salary offered and degree of education required. This dataset originally stems from the Employment Scam Aegean Dataset collected by~\citep{vidros2017automatic}: \url{https://www.kaggle.com/shivamb/real-or-fake-fake-jobposting-prediction}. The license of the original dataset: \textbf{CC0 Public Domain}.

\textbf{airbnb}: Predict the price label of AirBnb listings (in Melbourne, Australia) based on information from the listing page including various text descriptions and many numeric features (e.g. host's response-rate, number of bed/bath-rooms) and categorical features (e.g. property type, superhost or not). This dataset was released via the InsideAirbnb initiative: \url{https://www.kaggle.com/tylerx/melbourne-airbnb-open-data}. The license of the original dataset: \textbf{CC0 Public Domain}.

\textbf{channel}: Predict which news category (i.e. channel) a Mashable.com news article belongs to based on the text of its title, as well as auxiliary numerical features like the number of words in the article, its average token length, how many keywords are listed, etc. The original version of this dataset was collected by~\citep{fernandes2015proactive}: \url{https://archive.ics.uci.edu/ml/datasets/online+news+popularity}. The license of the original dataset: \textbf{CC BY 4.0}.

\textbf{qaa}: Given a question and an answer (from the Crowdsource team at Google) as well as an additional category feature, predict the (subjective) type of the answer in relation to the question. Representing a predominantly NLP task that requires deep language understanding (though the most accurate models must also consider the category), this dataset stems from a 2019 Kaggle competition: \url{https://www.kaggle.com/c/google-quest-challenge}. The license of the original dataset: Competition Use , Non-Commercial Purposes \& Academic Research, Commercial Purposes.

\textbf{qaq}: Given a question and an answer (from the Crowdsource team at Google) as well as additional category features, predict the (subjective) type of the question in relation to the answer. These data stem from the same source as qaa, where the different labels were both prediction targets in the original (multi-label) Kaggle competition.

\textbf{cloth}: Predict the score of a customer review of clothing items (sold by an anonymous retailer) based on the review text, how much positive feedback the review has received (numeric), and additional features about the product like its department (categorical). The data were collected by : \url{https://www.kaggle.com/nicapotato/womens-ecommerce-clothing-reviews}. The license of the original dataset: \textbf{CC0 Public Domain}.

\textbf{ptech}: Given a meme, identify if the smear persuasive technique is used both in the textual and visual content of the meme. This dataset originally stems from the third task of SemEval-2021 Task 6~\citep{SemEval2021:task6}: \url{https://github.com/di-dimitrov/SEMEVAL-2021-task6-corpus}. This was originally a multi-label classification task. However, we are now focusing solely on the ``smear'' label for binary classification. During data preprocessing, we combine the original training set with the validation set, removing duplicate samples to create a new training set. The original test set is retained for testing. The license of the original dataset: free for general research use.

\textbf{memotion}: Given an Internet meme, classify it as a positive, negative or neutral meme. This dataset original stems from the first task of ~\citep{chhavi2020memotion}. The license of the original dataset: is allowed to be used in any paper, only upon citation.

\textbf{food101}: Predict the food category (101 total) from the text of a recipe and the photo of the food. This dataset originally stems from \url{https://www.kaggle.com/datasets/gianmarco96/upmcfood101}. We randomly select 20\% of the original data from each category to create new training and test sets. The license of the original dataset: public. 

\textbf{aep}: Predict the action type (73 total) that causes an effect, based on a text description of the effect and an image depicting the effect. This dataset originally stems from~\citep{gao2018action}: \url{https://huggingface.co/datasets/sled-umich/Action-Effect}, containing language data and image data. The format of the language data for each line includes the action type, an effect sentence, and multiple effect phrases. The image data consists of images depicting the effects of actions. For each action type, an average of 15 positive images and 15 negative images were collected. In our benchmark, we randomly select effect sentences and images for each action type. The number of randomly selected samples for each action type is the minimum of the available effect sentences and images. The license of the original dataset: cite if used.

\textbf{fakeddit}: Predict the fake news type based on the provided news images and titles. This dataset originally stems from~\citep{nakamura2019r}: \url{https://github.com/entitize/Fakeddit}. The samples are labeled according to 2-way, 3-way, and 6-way classification categories. We use the most challenging 6-way categories in our benchmark. Additionally, we randomly select 3\% and 10\% of the original training set and test set respectively to create new training and test sets. The license of the original dataset: cite if used.

\textbf{ccd}: Predict whether ego-vehicles are involved in car accidents based on the accident images and metadata including environmental attributes (day/night, snowy/rainy/good weather conditions). This dataset originally stems from \url{https://www.kaggle.com/datasets/asefjamilajwad/car-crash-dataset-ccd}, which extracts frames of the videos from the CarCrashDataset~\citep{BaoMM2020} into images to be used for training. We use the image corresponding to the exact video frame when the accident occurred. We randomly split the original dataset at 3:1 ratio for new training set and test set.  The license of the original dataset: \textbf{MIT License}.

\textbf{HAM}: Predict pigmented skin lesions category (7 total) based on dermatoscopic images and metadata including basic information about patients (age, sex), how the lesions are confirmed and locations of the lesions. This dataset originally stems from~\citep{Tschandl2018_HAM10000}: \url{https://dataverse.harvard.edu/dataset.xhtml?persistentId=doi:10.7910/DVN/DBW86T}. We randomly split 20\% of the data from the original training set for testing, leaving the remainder as the training set. The license of the original dataset: \textbf{CC BY-NC 4.0 Deed}.

\textbf{wikiart}: Identify the artists (129 total) behind the artworks based on images and metadata including genres and styles. This dataset originally stems from \url{https://huggingface.co/datasets/huggan/wikiart}. We randomly select 25\% of the original dataset as a subset, then randomly choose 25\% of this subset for testing, leaving the remainder as the training set. The license of the original dataset: can be used only for non-commercial research purpose.

\textbf{cd18}: Predict the price of a cellphone using the images of its appearance and its specifications including weight, storage, battery type and so on. This dataset originally stems from~\citep{zehtab2023multimodal}: \url{https://github.com/AidinZe/CD18-Cellphone-Dataset-with-18-Features}. We randomly split the original dataset at 4:1 ratio for new training set and test set.  The license of the original dataset: cite if used.

\textbf{DVM}: Predict the selling price of cars based on car images and the metadata including fuel type, the number of seats, body type and so on. This dataset originally stems from~\citep{huang2022dvm}: \url{https://deepvisualmarketing.github.io}. The original dataset contains multiple metadata tables. We use Ad table as the metadata, which contains more than 0.25 million used car advertisements. We further select 25\% of the metadata randomly, and split the subset at 3:1 ratio for a new training set and test set. The original dataset provide car images with different views, and we use the front view. The license of the original dataset: \textbf{CC BY-NC 4.0}.

\textbf{petfinder}: Predict the speed at which a pet is adopted, based on the its image and metadata including text description, health situation, fur length, color and so on. This dataset originally stems from \url{https://www.kaggle.com/competitions/petfinder-adoption-prediction}. We randomly split the original training set at 4:1 ratio for new training and test sets. The license of the original dataset: for the purposes of the Competition, participation on Kaggle Website forums, academic research and education, and other non-commercial purposes.

\textbf{covid}: Predict the pneumonia category (25 total) based on the chestxray images and metadata including basic information of patients(age, sex, survival situation), clinical notes and so on. This dataset originally stems from~\citep{cohen2020covidProspective}: \url{https://github.com/ieee8023/covid-chestxray-dataset}. We randomly split the original dataset at 3:1 ratio for new training set and test set. The license of the original dataset: \textbf{Apache 2.0, CC BY-NC-SA 4.0, CC BY 4.0}.

\textbf{artm}: Predict the movement of an artwork (34 total) based on its image and metadata including title, price, condition and so on. This dataset originally stems from \url{https://www.kaggle.com/datasets/flkuhm/art-price-dataset}. We randomly split the data at 3:1 ratio for new training set and test set. The license of the original dataset: \textbf{CC BY-NC-SA 4.0}.

\textbf{seattle}: Predict the price of living in a hotel based on its image and metadata including text description, review from customers, room type and so on. This dataset originally stems from \url{https://www.kaggle.com/datasets/airbnb/seattle}. We randomly split the original training set at 4:1 ratio for new training and test sets. The license of the original dataset: \textbf{CC0 1.0}.

\textbf{goodreads}: Predict the rating of a book based on its cover image and metadata including title, content descriptions, genre and so on. This dataset originally stems from \url{https://www.kaggle.com/datasets/mdhamani/goodreads-books-100k}. We remove non-English sentences, randomly select 16\% of the original dataset and split this subset at 3:1 ratio for new training and test sets. The license of the original dataset: \textbf{CC0 1.0}.

\textbf{KARD}: Predict the rating of an anime based on its poster image and metadata including its name, genre, number of users who added this anime to their favorites and so on. This dataset originally stems from \url{https://www.kaggle.com/datasets/mersico/kagemura-anime-recommendation-dataset-kard}. We use the English metadata file it provides, and randomly split the metadata at 3:1 ratio for new training and test sets. The license of the original dataset: \textbf{GNU Affero General Public License}.

\section{Detailed Experiment Results}
\label{sec:detail_exp_res}

We used AWS EC2 p3dn.24xlarge equipped with 8 NVIDIA V100 GPUs for running experiments with LF-LLM model (LLM is the late-fusion module) and AWS EC2 g5.48xlarge instances equipped with 8 NVIDIA A10 GPUs for other experiment runs. The total GPU time used for training is about 6177.96 hours, which includes 3 repeats on 22 datasets. The following are the main experiment results for each dataset, along with supplementary experiments. These include comparisons among different serialization methods and simulations of various scenarios with different ratios of missing modalities in the training and test sets.

\subsection{Basic Tricks}

\begin{table}[H]\Huge
    \caption{Results of adding the basic tricks incrementally.}
    \centering
    \resizebox{\textwidth}{!}{%
    \renewcommand\arraystretch{1.5}
    \begin{tabular}{lccccccccccccccc}
        \toprule
         \multirow{2}*{\bf Method}  & \multicolumn{8}{c}{\bf Text+Tabular} & \multicolumn{7}{c}{\bf Image+Text} \\
         % & \multicolumn{7}{c}{\bf Image+Tabular} & \multicolumn{8}{c}{\bf Image+Text+Tabular} & \multirow{2}*{\bf avg $\uparrow$} & \multirow{2}*{\bf mrr $\uparrow$}

         \cmidrule(lr){2-9} \cmidrule(lr){10-16}  
         % \cmidrule(lr){17-23}  \cmidrule(lr){24-31} 

         & fake & airbnb & channel & qaa & qaq & cloth & \textbf{avg} & \textbf{mrr} & ptech & memotion & food101 & aep & fakeddit & \textbf{avg} & \textbf{mrr} \\
 
         % & roc\_auc & acc & acc & r2 & r2 & r2 &  \\
        \midrule
         Baseline & 0.738$_{0.052}$ & 0.406$_{0.004}$ & 0.521$_{0.002}$ & 0.420$_{0.033}$ & 0.458$_{0.035}$ & 0.606$_{0.180}$ & 0.525$_{0.039}$ & 0.376$_{0.017}$ & 0.556$_{0.017}$ & 0.592$_{0.000}$ & 0.717$_{0.003}$ & 0.542$_{0.015}$ & 0.867$_{0.005}$ & 0.655$_{0.007}$ & 0.420$_{0.038}$\\

        +Greedy Soup & 0.804$_{0.077}$ & 0.409$_{0.002}$ & 0.522$_{0.002}$ & 0.457$_{0.014}$ & 0.462$_{0.011}$ & 0.613$_{0.185}$ & 0.545$_{0.039}$ & 0.381$_{0.028}$  & 0.555$_{0.016}$ & 0.592$_{0.000}$ & 0.791$_{0.067}$ & 0.544$_{0.009}$ & 0.867$_{0.005}$ & 0.670$_{0.019}$ & 0.383$_{0.036}$\\
        
        +Gradient Clip & 0.950$_{0.019}$ & 0.403$_{0.013}$ & 0.511$_{0.006}$ & 0.480$_{0.008}$ & 0.474$_{0.004}$ & 0.747$_{0.001}$ & 0.594$_{0.002}$ & 0.492$_{0.136}$ & 0.573$_{0.012}$ & 0.592$_{0.000}$ & 0.839$_{0.089}$ & 0.564$_{0.000}$ & 0.870$_{0.005}$ & 0.688$_{0.021}$ & 0.669$_{0.143}$\\

        +LR Warmup & 0.964$_{0.006}$ & 0.405$_{0.004}$ & 0.510$_{0.005}$ & 0.473$_{0.007}$ & 0.480$_{0.007}$ & 0.736$_{0.006}$ & 0.595$_{0.002}$ & 0.469$_{0.106}$ & 0.568$_{0.018}$ & 0.585$_{0.006}$ & 0.928$_{0.001}$ & 0.569$_{0.003}$ & 0.877$_{0.005}$ & \textbf{0.705$_{0.006}$} & \textbf{0.727$_{0.233}$} \\

        +Layerwise Decay & 0.956$_{0.009}$ & 0.417$_{0.002}$ & 0.511$_{0.004}$ & 0.480$_{0.004}$ & 0.483$_{0.009}$ & 0.752$_{0.006}$ & \textbf{0.600$_{0.003}$} & \textbf{0.599$_{0.145}$} & 0.556$_{0.006}$ & 0.586$_{0.004}$ & 0.928$_{0.002}$ & 0.557$_{0.005}$ & 0.879$_{0.002}$ & 0.701$_{0.003}$ & 0.570$_{0.099}$ \\

        \bottomrule
    \end{tabular}
    }

    \resizebox{\textwidth}{!}{%
    \renewcommand\arraystretch{1.5}
    \begin{tabular}{lccccccccccccccccc}
        \toprule
         \multirow{2}*{\bf Method} & \multicolumn{7}{c}{\bf Image+Tabular} & \multicolumn{8}{c}{\bf Image+Text+Tabular} & \multirow{2}*{\bf avg $\uparrow$} & \multirow{2}*{\bf mrr $\uparrow$}\\

         \cmidrule(lr){2-8} \cmidrule(lr){9-16}

         & ccd & HAM & wikiart & cd18 & DVM & \textbf{avg} & \textbf{mrr} & petfinder & covid & artm & seattle & goodreads & KARD & \textbf{avg} & \textbf{mrr}  \\

         \midrule

        Baseline & 0.914$_{0.007}$ & 0.913$_{0.007}$ & 0.740$_{0.003}$ & 0.539$_{0.003}$ & 0.917$_{0.000}$ & 0.805$_{0.000}$ & 0.272$_{0.032}$ & 0.353$_{0.014}$ & 0.829$_{0.017}$ & 0.608$_{0.027}$ & 0.478$_{0.038}$ & 0.050$_{0.008}$ & 0.770$_{0.011}$ & 0.515$_{0.011}$ & 0.280$_{0.058}$  & 0.615$_{0.014}$ & 0.336$_{0.014}$ \\

        +Greedy Soup & 0.914$_{0.007}$ & 0.921$_{0.001}$ & 0.744$_{0.000}$ & 0.539$_{0.002}$ & 0.917$_{0.000}$ & 0.807$_{0.002}$ & 0.308$_{0.007}$  & 0.383$_{0.016}$ & 0.839$_{0.004}$ & 0.593$_{0.040}$ & 0.483$_{0.042}$ & 0.085$_{0.031}$ & 0.771$_{0.012}$ & 0.526$_{0.009}$ & 0.320$_{0.051}$  & 0.628$_{0.015}$ & 0.348$_{0.005}$ \\
        
        +Gradient Clip & 0.912$_{0.008}$ & 0.920$_{0.002}$ & 0.748$_{0.003}$ & 0.591$_{0.009}$ & 0.932$_{0.002}$ & 0.821$_{0.002}$ & 0.354$_{0.009}$ & 0.381$_{0.020}$ & 0.874$_{0.011}$ & 0.593$_{0.012}$ & 0.601$_{0.010}$ & 0.125$_{0.047}$ & 0.795$_{0.005}$ & 0.562$_{0.005}$ & 0.411$_{0.034}$  & 0.658$_{0.004}$ & 0.479$_{0.034}$ \\

        +LR Warmup & 0.915$_{0.003}$ & 0.922$_{0.003}$ & 0.755$_{0.004}$ & 0.618$_{0.010}$ & 0.934$_{0.000}$ & 0.829$_{0.001}$ & \textbf{0.763$_{0.107}$}  & 0.379$_{0.015}$ & 0.874$_{0.004}$ & 0.597$_{0.028}$ & 0.634$_{0.009}$ & 0.239$_{0.018}$ & 0.802$_{0.004}$ & 0.588$_{0.005}$ & 0.645$_{0.017}$ & 0.671$_{0.001}$ & \textbf{0.642$_{0.028}$} \\

        +Layerwise Decay & 0.916$_{0.008}$ & 0.926$_{0.003}$ & 0.757$_{0.002}$ & 0.617$_{0.007}$ & 0.932$_{0.002}$ & \textbf{0.830$_{0.004}$} & 0.717$_{0.062}$ & 0.399$_{0.002}$ & 0.865$_{0.011}$ & 0.603$_{0.010}$ & 0.624$_{0.017}$ & 0.244$_{0.009}$ & 0.804$_{0.007}$ & \textbf{0.590$_{0.003}$} & \textbf{0.648$_{0.052}$} & \textbf{0.672$_{0.002}$} & 0.633$_{0.089}$ \\
       
         % & roc\_auc & acc & acc & r2 & r2 & r2 &  \\
        \midrule
        
        \bottomrule
    \end{tabular}
    }
    
    \label{tab:basic_detail}
\end{table}

\subsection{Multimodal Fusion Strategies}

% How to effectively combining different modalities is a key design challenge. The fusion method must be scalable to handle arbitrary combinations of images, text, and tabular data.

\begin{table}[H]\Huge
    \caption{Results of  multimodal fusion strategies. }
    \centering
    \resizebox{\textwidth}{!}{%
    \renewcommand\arraystretch{1.5}
    \begin{tabular}{lccccccccccccccc}
        \toprule
         \multirow{2}*{\bf Method}  & \multicolumn{8}{c}{\bf Text+Tabular} & \multicolumn{7}{c}{\bf Image+Text} \\
         % & \multicolumn{7}{c}{\bf Image+Tabular} & \multicolumn{8}{c}{\bf Image+Text+Tabular} & \multirow{2}*{\bf avg $\uparrow$} & \multirow{2}*{\bf mrr $\uparrow$}

         \cmidrule(lr){2-9} \cmidrule(lr){10-16}  
         % \cmidrule(lr){17-23}  \cmidrule(lr){24-31} 

         & fake & airbnb & channel & qaa & qaq & cloth & \textbf{avg} & \textbf{mrr} & ptech & memotion & food101 & aep & fakeddit & \textbf{avg} & \textbf{mrr} \\
         \midrule
 
          LF-MLP(Baseline+) & 0.956$_{0.009}$ & 0.417$_{0.002}$ & 0.511$_{0.004}$ & 0.480$_{0.004}$ & 0.483$_{0.009}$ & 0.752$_{0.006}$ & 0.600$_{0.003}$ & 0.597$_{0.049}$ & 0.556$_{0.006}$ & 0.586$_{0.004}$ & 0.928$_{0.002}$ & 0.557$_{0.005}$ & 0.879$_{0.002}$ & 0.701$_{0.003}$ & 0.399$_{0.043}$ \\
         
         LF-Transformer & 0.959$_{0.011}$ & 0.420$_{0.002}$ & 0.513$_{0.007}$ & 0.470$_{0.006}$ & 0.491$_{0.014}$ & 0.754$_{0.003}$ & \textbf{0.601$_{0.001}$} & \textbf{0.644$_{0.062}$} & 0.568$_{0.028}$ & 0.591$_{0.001}$ & 0.892$_{0.007}$ & 0.597$_{0.008}$ & 0.872$_{0.001}$ & 0.704$_{0.007}$ & 0.354$_{0.063}$ \\
         
         LF-Aligned & 0.945$_{0.014}$ & 0.399$_{0.003}$ & 0.482$_{0.006}$ & 0.362$_{0.027}$ & 0.343$_{0.018}$ & 0.695$_{0.002}$ & 0.538$_{0.005}$ & 0.240$_{0.027}$ & 0.728$_{0.014}$ & 0.578$_{0.015}$ & 0.937$_{0.006}$ & 0.718$_{0.013}$ & 0.910$_{0.002}$ & \textbf{0.774$_{0.002}$} & \textbf{0.891$_{0.077}$} \\

         LF-LLM & 0.939$_{0.003}$ & 0.407$_{0.007}$ & 0.506$_{0.002}$ & 0.332$_{0.082}$ & 0.122$_{0.090}$ & 0.474$_{0.107}$ & 0.463$_{0.003}$ & 0.261$_{0.059}$ & 0.623$_{0.012}$ & 0.592$_{0.000}$ & 0.880$_{0.021}$ & 0.519$_{0.024}$ & 0.875$_{0.000}$ & 0.698$_{0.009}$ & 0.389$_{0.057}$  \\

         LF-SF & 0.972$_{0.004}$ & 0.424$_{0.002}$ & 0.475$_{0.028}$ & 0.455$_{0.008}$ & 0.448$_{0.021}$ & 0.735$_{0.005}$ & 0.585$_{0.008}$ & 0.544$_{0.014}$ & 0.557$_{0.042}$ & 0.592$_{0.000}$ & 0.267$_{0.364}$ & 0.006$_{0.005}$ & 0.713$_{0.014}$ & 0.427$_{0.072}$ & 0.322$_{0.042}$ \\

         Early-fusion &  0.921$_{0.006}$ & 0.408$_{0.009}$ & 0.486$_{0.009}$ & 0.205$_{0.049}$ & 0.255$_{0.040}$ & 0.633$_{0.010}$ & 0.484$_{0.012}$ & 0.195$_{0.012}$ & 0.565$_{0.029}$ & 0.592$_{0.000}$ & 0.862$_{0.002}$ & 0.124$_{0.022}$ & 0.752$_{0.006}$ & 0.579$_{0.002}$ & 0.334$_{0.057}$ \\

        \bottomrule
    \end{tabular}
    }

    \resizebox{\textwidth}{!}{%
    \renewcommand\arraystretch{1.5}
    \begin{tabular}{lccccccccccccccccc}
        \toprule
         \multirow{2}*{\bf Method} & \multicolumn{7}{c}{\bf Image+Tabular} & \multicolumn{8}{c}{\bf Image+Text+Tabular} & \multirow{2}*{\bf avg $\uparrow$} & \multirow{2}*{\bf mrr $\uparrow$}\\

         \cmidrule(lr){2-8} \cmidrule(lr){9-16}

         & ccd & HAM & wikiart & cd18 & DVM & \textbf{avg} & \textbf{mrr} & petfinder & covid & artm & seattle & goodreads & KARD & \textbf{avg} & \textbf{mrr}  \\

         \midrule

         LF-MLP(Baseline+) & 0.916$_{0.008}$ & 0.926$_{0.003}$ & 0.757$_{0.002}$ & 0.617$_{0.007}$ & 0.932$_{0.002}$ & 0.830$_{0.004}$ & 0.374$_{0.069}$ & 0.399$_{0.002}$ & 0.865$_{0.011}$ & 0.603$_{0.010}$ & 0.624$_{0.017}$ & 0.244$_{0.009}$ & 0.804$_{0.007}$ & 0.590$_{0.003}$ & 0.546$_{0.047}$ & 0.672$_{0.002}$ & 0.488$_{0.044}$ \\
         
         LF-Transformer & 0.918$_{0.009}$ & 0.921$_{0.009}$ & 0.772$_{0.004}$ & 0.593$_{0.022}$ & 0.934$_{0.001}$ & 0.828$_{0.002}$ & 0.392$_{0.083}$ & 0.386$_{0.007}$ & 0.832$_{0.016}$ & 0.601$_{0.018}$ & 0.593$_{0.013}$ & 0.156$_{0.071}$ & 0.792$_{0.006}$ & 0.560$_{0.012}$ & 0.331$_{0.037}$  & 0.665$_{0.002}$ & 0.435$_{0.033}$\\
         
         LF-Aligned & 0.950$_{0.005}$ & 0.923$_{0.003}$ & 0.787$_{0.004}$ & 0.645$_{0.011}$ & 0.938$_{0.001}$ & \textbf{0.848$_{0.004}$} & \textbf{0.839$_{0.064}$} & 0.413$_{0.004}$ & 0.935$_{0.005}$ & 0.637$_{0.022}$ & 0.695$_{0.005}$ & 0.261$_{0.004}$ & 0.754$_{0.034}$ & \textbf{0.616$_{0.008}$} & \textbf{0.870$_{0.007}$}  & \textbf{0.683$_{0.001}$} & \textbf{0.696$_{0.033}$} \\

         LF-LLM & 0.917$_{0.006}$ & 0.926$_{0.001}$ & 0.752$_{0.011}$ & 0.603$_{0.019}$ & 0.934$_{0.001}$ & 0.826$_{0.002}$ & 0.411$_{0.016}$ & 0.323$_{0.030}$ & 0.859$_{0.038}$ & 0.561$_{0.023}$ & 0.672$_{0.006}$ & 0.128$_{0.036}$ & 0.796$_{0.006}$ & 0.556$_{0.008}$ & 0.361$_{0.020}$  & 0.625$_{0.004}$ & 0.352$_{0.019}$ \\

         LF-SF & 0.539$_{0.013}$ & 0.711$_{0.003}$ & 0.547$_{0.002}$ & 0.606$_{0.021}$ & 0.922$_{0.000}$ & 0.665$_{0.007}$ & 0.208$_{0.024}$ &  0.000$_{0.000}$ & 0.631$_{0.000}$ & 0.258$_{0.009}$ & 0.126$_{0.179}$ & 0.001$_{0.001}$ & 0.753$_{0.005}$ & 0.295$_{0.030}$ & 0.174$_{0.003}$ & 0.488$_{0.016}$ & 0.316$_{0.001}$\\

         Early-fusion & 0.698$_{0.161}$ & 0.864$_{0.038}$ & 0.693$_{0.045}$ & 0.552$_{0.030}$ & 0.926$_{0.012}$ & 0.747$_{0.057}$ & 0.273$_{0.118}$ & 0.275$_{0.017}$ & 0.649$_{0.025}$ & 0.311$_{0.037}$ & 0.546$_{0.020}$ & 0.136$_{0.014}$ & 0.748$_{0.028}$ & 0.444$_{0.007}$ & 0.208$_{0.007}$  & 0.555$_{0.015}$ & 0.248$_{0.040}$\\

        \bottomrule
    \end{tabular}
    }
    
    \label{tab:struc_detail}
\end{table}

\begin{table}[H]
    \caption{Trainable parameter number (M) comparison among different multimodal fusion strategies. }
    \centering
    \resizebox{0.6\textwidth}{!}{%
    \renewcommand\arraystretch{1.5}
    \begin{tabular}{lcccc}
        \toprule
         {\bf Fusion Model} & {\bf Text+Tabular} &  {\bf Image+Text} &  {\bf Image+Tabular} &  {\bf Image+Text+Tabular}\\
         \midrule
         LF-MLP(Baseline+)  & 185&383& 199& 384\\
         LF-Transformer & 193 & 413 & 229& 414\\
         LF-Aligned & 126 & 428 & 306 & 429\\
         LF-LLM & 190 & 390 & 205 & 392\\
         LF-SF & 184 & 381 & 196 & 381 \\
         Early Fusion & 402 & 402 & 302 & 402\\

        \bottomrule
    \end{tabular}
    }

    \label{tab:struc_cost}
\end{table}

\subsection{Converting Tabular Data into Text}

\begin{table}[H]
    \caption{Results of converting tabular data into text data on \textit{datasets with tabular data}.}
    \centering

    \begingroup
    \Huge
    \resizebox{\textwidth}{!}{%
    \renewcommand\arraystretch{1.5}
    \begin{tabular}{lccccccccccccccc}
        \toprule
         \multirow{2}*{\bf Method.} &  \multicolumn{8}{c}{\bf Text+Tabular} &  \multicolumn{7}{c}{\bf Image+Tabular} \\

         \cmidrule(lr){2-9}  \cmidrule(lr){10-16}
         
         & fake & airbnb & channel & qaa & qaq & cloth  & \textbf{avg} & \textbf{mrr} & ccd & HAM & wikiart & cd18 & DVM & \textbf{avg} & \textbf{mrr}  \\
 
         % & roc\_auc & acc & acc & r2 & r2 & r2 &  \\
        \midrule
         Baseline+ & 0.956$_{0.009}$ & 0.417$_{0.002}$ & 0.511$_{0.004}$ & 0.480$_{0.004}$ & 0.483$_{0.009}$ & 0.752$_{0.006}$ & 0.600$_{0.003}$ & 0.713$_{0.151}$ & 0.916$_{0.008}$ & 0.926$_{0.003}$ & 0.757$_{0.002}$ & 0.617$_{0.007}$ & 0.932$_{0.002}$ & 0.830$_{0.004}$ & 0.689$_{0.177}$ \\
         
        Convert Categorical & 0.968$_{0.006}$ & 0.416$_{0.006}$ & 0.513$_{0.0}$ & 0.481$_{0.011}$ & 0.485$_{0.009}$ & 0.751$_{0.002}$ & \textbf{0.602$_{0.001}$} & \textbf{0.731$_{0.073}$} & 0.925$_{0.004}$ & 0.925$_{0.003}$ & 0.755$_{0.002}$ & 0.614$_{0.007}$ & 0.967$_{0.0}$ & \textbf{0.837$_{0.001}$} & \textbf{0.800$_{0.216}$}  \\
        
         Convert Numeric & 0.956$_{0.009}$ & 0.410$_{0.003}$ & 0.499$_{0.005}$ & 0.480$_{0.004}$ & 0.483$_{0.009}$ & 0.753$_{0.002}$ & 0.597$_{0.001}$ & 0.611$_{0.068}$ & 0.916$_{0.008}$ & 0.925$_{0.002}$ & 0.757$_{0.002}$ & 0.498$_{0.005}$ & 0.932$_{0.002}$ & 0.806$_{0.003}$ & 0.522$_{0.103}$ \\
         
        \bottomrule
    \end{tabular}
    }
    \endgroup

    \resizebox{\textwidth}{!}{%
    \renewcommand\arraystretch{1.5}
    \begin{tabular}{lcccccccccc}
        \toprule
         \multirow{2}*{\bf Method.} & \multicolumn{8}{c}{\bf Image+Text+Tabular} & \multirow{2}*{\bf avg $\uparrow$} & \multirow{2}*{\bf mrr $\uparrow$}\\

         \cmidrule(lr){2-9} 
         
         & petfinder & covid & artm & seattle & goodreads & KARD  & \textbf{avg} & \textbf{mrr} \\

         % & roc\_auc & acc & acc & r2 & r2 & r2 &  \\
        \midrule
         Baseline+ & 0.399$_{0.002}$ & 0.865$_{0.011}$ & 0.603$_{0.01}$ & 0.624$_{0.017}$ & 0.244$_{0.009}$ & 0.804$_{0.007}$ & \textbf{0.590$_{0.003}$} & \textbf{0.713$_{0.069}$} & 0.664$_{0.002}$ & 0.706$_{0.097}$ \\
         
        Convert Categorical & 0.385$_{0.01}$ & 0.853$_{0.022}$ & 0.580$_{0.014}$ & 0.629$_{0.010}$ & 0.242$_{0.006}$ & 0.819$_{0.004}$ & 0.585$_{0.006}$ & 0.602$_{0.052}$  & \textbf{0.665$_{0.002}$} & \textbf{0.706$_{0.081}$} \\
        
         Convert Numeric & 0.368$_{0.003}$ & 0.847$_{0.010}$ & 0.590$_{0.012}$ & 0.556$_{0.010}$ & 0.238$_{0.007}$ & 0.875$_{0.002}$ & 0.579$_{0.004}$ & 0.565$_{0.026}$ & 0.652$_{0.001}$ & 0.569$_{0.037}$\\
         
        \bottomrule
    \end{tabular}
    }

    \label{tab:covert_tabular_detail}
\end{table}

We also investigate some serialization methods mentioned in ~\citep{carballo2022tabtext,hegselmann2023tabllm,nam2023semi,wang2023unipredict,jaitly2023towards} when converting categorical data to text. Consider a cell from a column named ``C'' that has value X, the methods can be categorized into: 1) Direct: X -> ``X'', changing the value into a text string directly. 2) List: X -> ``C: X'', using a list template for transformation. 3) Text: ``C is X'', using a text template for transformation. 4) Latex: X -> ``X \& '', using a latex template for transformation.

\begin{table}[H]
    \caption{Results of different serialization methods on \textit{datasets with tabular data}.}
    \centering

    \begingroup
    \Huge
    \resizebox{\textwidth}{!}{%
    \renewcommand\arraystretch{1.5}
    \begin{tabular}{lccccccccccccccc}
        \toprule
         \multirow{2}*{\bf Method.} &  \multicolumn{8}{c}{\bf Text+Tabular} &  \multicolumn{7}{c}{\bf Image+Tabular} \\

          \cmidrule(lr){2-9}  \cmidrule(lr){10-16}
          
         & fake & airbnb & channel & qaa & qaq & cloth  & \textbf{avg} & \textbf{mrr} & ccd & HAM & wikiart & cd18 & DVM & \textbf{avg} & \textbf{mrr}  \\
 
         % & roc\_auc & acc & acc & r2 & r2 & r2 &  \\
        \midrule

         Baseline+ & 0.956$_{0.009}$ & 0.417$_{0.002}$ & 0.511$_{0.004}$ & 0.480$_{0.004}$ & 0.483$_{0.009}$ & 0.752$_{0.006}$ & 0.600$_{0.003}$ & 0.373$_{0.096}$ & 0.916$_{0.008}$ & 0.926$_{0.003}$ & 0.757$_{0.002}$ & 0.617$_{0.007}$ & 0.932$_{0.002}$ & 0.830$_{0.004}$ & 0.469$_{0.207}$\\
         
        Direct & 0.976$_{0.003}$ & 0.421$_{0.001}$ & 0.509$_{0.005}$ & 0.473$_{0.012}$ & 0.494$_{0.007}$ & 0.748$_{0.002}$ & \textbf{0.604$_{0.001}$} & \textbf{0.556$_{0.140}$} & 0.923$_{0.001}$ & 0.924$_{0.002}$ & 0.757$_{0.001}$ & 0.622$_{0.006}$ & 0.968$_{0.001}$ & \textbf{0.839$_{0.001}$} & \textbf{0.622$_{0.102}$} \\

        List & 0.973$_{0.002}$ & 0.397$_{0.008}$ & 0.518$_{0.006}$ & 0.480$_{0.015}$ & 0.488$_{0.011}$ & 0.747$_{0.006}$ & 0.600$_{0.002}$ & 0.506$_{0.023}$ & 0.924$_{0.005}$ & 0.927$_{0.002}$ & 0.752$_{0.005}$ & 0.614$_{0.004}$ & 0.963$_{0.003}$ & 0.836$_{0.001}$ & 0.474$_{0.174}$\\
        
        Text & 0.975$_{0.001}$ & 0.398$_{0.009}$ & 0.515$_{0.005}$ & 0.479$_{0.012}$ & 0.483$_{0.014}$ & 0.756$_{0.000}$ & 0.601$_{0.003}$ & 0.498$_{0.053}$ & 0.922$_{0.004}$ & 0.925$_{0.002}$ & 0.755$_{0.001}$ & 0.604$_{0.004}$ & 0.965$_{0.000}$ & 0.834$_{0.001}$ & 0.331$_{0.066}$\\

        Latex & 0.968$_{0.006}$ & 0.416$_{0.006}$ & 0.513$_{0.000}$ & 0.481$_{0.011}$ & 0.485$_{0.009}$ & 0.751$_{0.002}$ & 0.602$_{0.001}$ & 0.475$_{0.028}$ & 0.925$_{0.004}$ & 0.925$_{0.003}$ & 0.755$_{0.002}$ & 0.614$_{0.007}$ & 0.967$_{0.000}$ & 0.837$_{0.001}$ & 0.463$_{0.041}$ \\

        \bottomrule
    \end{tabular}
    }
    \endgroup

    \resizebox{\textwidth}{!}{%
    \renewcommand\arraystretch{1.2}
    \begin{tabular}{lcccccccccc}
        \toprule
         \multirow{2}*{\bf Method.} & \multicolumn{8}{c}{\bf Image+Text+Tabular} & \multirow{2}*{\bf avg $\uparrow$} & \multirow{2}*{\bf mrr $\uparrow$}\\

          \cmidrule(lr){2-9} 
          
         & petfinder & covid & artm & seattle & goodreads & KARD  & \textbf{avg} & \textbf{mrr} \\
 
         % & roc\_auc & acc & acc & r2 & r2 & r2 &  \\
        \midrule

         Baseline+ &  0.399$_{0.002}$ & 0.865$_{0.011}$ & 0.603$_{0.01}$ & 0.624$_{0.017}$ & 0.244$_{0.009}$ & 0.804$_{0.007}$ & \textbf{0.590$_{0.003}$} & \textbf{0.656$_{0.094}$}  & 0.664$_{0.002}$ & 0.501$_{0.103}$\\

        Direct & 0.377$_{0.011}$ & 0.827$_{0.008}$ & 0.584$_{0.012}$ & 0.595$_{0.011}$ & 0.246$_{0.002}$ & 0.814$_{0.001}$ & 0.574$_{0.001}$ & 0.386$_{0.055}$ & 0.662$_{0.001}$ & \textbf{0.516$_{0.099}$} \\

        List & 0.371$_{0.010}$ & 0.809$_{0.024}$ & 0.58$_{0.029}$ & 0.554$_{0.016}$ & 0.243$_{0.006}$ & 0.812$_{0.004}$ & 0.561$_{0.003}$ & 0.303$_{0.049}$ & 0.656$_{0.001}$ & 0.425$_{0.051}$  \\
        
        Text & 0.353$_{0.007}$ & 0.841$_{0.017}$ & 0.592$_{0.010}$ & 0.569$_{0.015}$ & 0.241$_{0.012}$ & 0.816$_{0.008}$ & 0.569$_{0.002}$ & 0.401$_{0.125}$ & 0.658$_{0.002}$ & 0.415$_{0.053}$\\

        Latex & 0.385$_{0.010}$ & 0.853$_{0.022}$ & 0.580$_{0.014}$ & 0.629$_{0.010}$ & 0.242$_{0.006}$ & 0.819$_{0.004}$ & 0.585$_{0.006}$ & 0.567$_{0.053}$ & \textbf{0.665$_{0.002}$} & 0.504$_{0.030}$ \\

        \bottomrule
    \end{tabular}
    }
    
    \label{tab:serial_methods}
\end{table}

In \Cref{tab:serial_methods}, all the serialization methods improve model performance in text+tabular and image+tabular compositions, while fall behind the baseline in image+text+tabular composition. Since all methods exhibit similar patterns, and using a latex template outperforms the baseline in both average value and mrr value across all datasets, we set it as the default method for converting categorical data into text.

\subsection{Cross-modal Alignment}

\begin{table}[H]\Huge
    \caption{Results of cross-modality alignment.}
    \centering
    \resizebox{\textwidth}{!}{%
    \renewcommand\arraystretch{1.5}
    \begin{tabular}{lccccccccccccccc}
        \toprule
         \multirow{2}*{\bf Method}  & \multicolumn{8}{c}{\bf Text+Tabular} & \multicolumn{7}{c}{\bf Image+Text} \\
         % & \multicolumn{7}{c}{\bf Image+Tabular} & \multicolumn{8}{c}{\bf Image+Text+Tabular} & \multirow{2}*{\bf avg $\uparrow$} & \multirow{2}*{\bf mrr $\uparrow$}

         \cmidrule(lr){2-9} \cmidrule(lr){10-16}  
         % \cmidrule(lr){17-23}  \cmidrule(lr){24-31} 

         & fake & airbnb & channel & qaa & qaq & cloth & \textbf{avg} & \textbf{mrr} & ptech & memotion & food101 & aep & fakeddit & \textbf{avg} & \textbf{mrr} \\
 
         % & roc\_auc & acc & acc & r2 & r2 & r2 &  \\
        \midrule
         Baseline+ & 0.956$_{0.009}$ & 0.417$_{0.002}$ & 0.511$_{0.004}$ & 0.480$_{0.004}$ & 0.483$_{0.009}$ & 0.752$_{0.006}$ & \textbf{0.600$_{0.003}$} & 0.648$_{0.170}$ & 0.556$_{0.006}$ & 0.586$_{0.004}$ & 0.928$_{0.002}$ & 0.557$_{0.005}$ & 0.879$_{0.002}$ & 0.701$_{0.003}$ & 0.422$_{0.079}$ \\
         
         % +Contra. Loss & 0.964 & 0.417 & 0.505 & 0.478 & 0.479 & 0.753 & 0.566 & 0.590 & 0.929 & 0.589 & 0.883 & 0.920 & 0.918 & 0.748 & 0.571 & 0.925 & 0.403 & 0.836 & 0.616 & 0.597 & 0.243 & 0.789 & 0.669  \\
         Positive-only & 0.974$_{0.002}$ & 0.408$_{0.007}$ & 0.505$_{0.007}$ & 0.475$_{0.008}$ & 0.479$_{0.007}$ & 0.750$_{0.002}$ & 0.599$_{0.002}$ & \textbf{0.657$_{0.035}$} & 0.560$_{0.023}$ & 0.583$_{0.013}$ & 0.932$_{0.003}$ & 0.576$_{0.005}$ & 0.884$_{0.003}$ & 0.707$_{0.007}$ & 0.711$_{0.042}$  \\
         
        Positive+Negative  & 0.963$_{0.004}$ & 0.418$_{0.005}$ & 0.502$_{0.006}$ & 0.476$_{0.012}$ & 0.491$_{0.010}$ & 0.751$_{0.003}$ & \textbf{0.600$_{0.002}$} & 0.611$_{0.138}$ & 0.591$_{0.012}$ & 0.592$_{0.000}$ & 0.929$_{0.001}$ & 0.605$_{0.016}$ & 0.884$_{0.004}$ & \textbf{0.720$_{0.007}$} & \textbf{0.856$_{0.042}$} \\
       
        \bottomrule
    \end{tabular}
    }
    \resizebox{\textwidth}{!}{%
    \renewcommand\arraystretch{1.5}
    \begin{tabular}{lccccccccccccccccc}
        \toprule
         \multirow{2}*{\bf Method} & \multicolumn{7}{c}{\bf Image+Tabular} & \multicolumn{8}{c}{\bf Image+Text+Tabular} & \multirow{2}*{\bf avg $\uparrow$} & \multirow{2}*{\bf mrr $\uparrow$}\\

         \cmidrule(lr){2-8} \cmidrule(lr){9-16}

         & ccd & HAM & wikiart & cd18 & DVM & \textbf{avg} & \textbf{mrr} & petfinder & covid & artm & seattle & goodreads & KARD & \textbf{avg} & \textbf{mrr}  \\

         \midrule

         Baseline+ & 0.916$_{0.008}$ & 0.926$_{0.003}$ & 0.757$_{0.002}$ & 0.617$_{0.007}$ & 0.932$_{0.002}$ & 0.830$_{0.004}$ & 0.611$_{0.140}$ & 0.399$_{0.002}$ & 0.865$_{0.011}$ & 0.603$_{0.010}$ & 0.624$_{0.017}$ & 0.244$_{0.009}$ & 0.804$_{0.007}$ & 0.590$_{0.003}$ & 0.556$_{0.023}$  & 0.672$_{0.002}$ & 0.563$_{0.090}$\\

        Positive-only & 0.923$_{0.007}$ & 0.923$_{0.002}$ & 0.753$_{0.004}$ & 0.621$_{0.008}$ & 0.934$_{0.001}$ & \textbf{0.831$_{0.002}$} & \textbf{0.778$_{0.096}$} & 0.412$_{0.002}$ & 0.884$_{0.018}$ & 0.614$_{0.017}$ & 0.655$_{0.015}$ & 0.247$_{0.009}$ & 0.802$_{0.011}$ & \textbf{0.602$_{0.007}$} & \textbf{0.824$_{0.013}$} & \textbf{0.677$_{0.002}$} & \textbf{0.742$_{0.031}$} \\

        Positive+Negative & 0.927$_{0.003}$ & 0.924$_{0.001}$ & 0.749$_{0.002}$ & 0.590$_{0.008}$ & 0.924$_{0.001}$ & 0.823$_{0.002}$ & 0.478$_{0.031}$ & 0.397$_{0.008}$ & 0.845$_{0.005}$ & 0.614$_{0.010}$ & 0.600$_{0.013}$ & 0.235$_{0.008}$ & 0.793$_{0.004}$ & 0.581$_{0.002}$ & 0.481$_{0.057}$  & 0.673$_{0.002}$ & 0.601$_{0.029}$ \\
        \bottomrule
    \end{tabular}
    }
    \label{tab:align_detail}
\end{table}

\subsection{Multimodal Data Augmentation}

\begin{table}[H]\Huge
    \caption{Results of multimodal data augmentation.}
    \centering
    \resizebox{\textwidth}{!}{%
    \renewcommand\arraystretch{1.5}
    \begin{tabular}{lccccccccccccccc}
        \toprule
         \multirow{2}*{\bf Method}  & \multicolumn{8}{c}{\bf Text+Tabular} & \multicolumn{7}{c}{\bf Image+Text} \\
         % & \multicolumn{7}{c}{\bf Image+Tabular} & \multicolumn{8}{c}{\bf Image+Text+Tabular} & \multirow{2}*{\bf avg $\uparrow$} & \multirow{2}*{\bf mrr $\uparrow$}

         \cmidrule(lr){2-9} \cmidrule(lr){10-16}  
         % \cmidrule(lr){17-23}  \cmidrule(lr){24-31} 

         & fake & airbnb & channel & qaa & qaq & cloth & \textbf{avg} & \textbf{mrr} & ptech & memotion & food101 & aep & fakeddit & \textbf{avg} & \textbf{mrr} \\
         \midrule
 
          Baseline+ & 0.956$_{0.009}$ & 0.417$_{0.002}$ & 0.511$_{0.004}$ & 0.480$_{0.004}$ & 0.483$_{0.009}$ & 0.752$_{0.006}$ & 0.600$_{0.003}$ & 0.495$_{0.195}$ & 0.556$_{0.006}$ & 0.586$_{0.004}$ & 0.928$_{0.002}$ & 0.557$_{0.005}$ & 0.879$_{0.002}$ & 0.701$_{0.003}$ & 0.417$_{0.131}$ \\

          Input Aug. & 0.962$_{0.002}$ & 0.419$_{0.004}$ & 0.509$_{0.002}$ & 0.475$_{0.012}$ & 0.493$_{0.006}$ & 0.749$_{0.001}$ & 0.601$_{0.003}$ & 0.426$_{0.064}$ & 0.579$_{0.023}$ & 0.581$_{0.010}$ & 0.933$_{0.002}$ & 0.580$_{0.012}$ & 0.879$_{0.000}$ &\textbf{ 0.710$_{0.004}$} & \textbf{0.683$_{0.049}$} \\

          Feature Aug.(Inde.) & 0.965$_{0.006}$ & 0.412$_{0.009}$ & 0.513$_{0.007}$ & 0.478$_{0.001}$ & 0.485$_{0.007}$ & 0.753$_{0.001}$ & 0.601$_{0.002}$ & 0.426$_{0.080}$ & 0.587$_{0.008}$ & 0.589$_{0.003}$ & 0.923$_{0.006}$ & 0.551$_{0.016}$ & 0.880$_{0.002}$ & 0.706$_{0.006}$ & 0.594$_{0.102}$ \\

          Feature Aug.(Joint) & 0.965$_{0.009}$ & 0.424$_{0.003}$ & 0.507$_{0.006}$ & 0.481$_{0.012}$ & 0.488$_{0.007}$ & 0.756$_{0.000}$ & \textbf{0.604$_{0.003}$} & \textbf{0.741$_{0.125}$} & 0.581$_{0.036}$ & 0.582$_{0.004}$ & 0.930$_{0.004}$ & 0.554$_{0.009}$ & 0.877$_{0.001}$ & 0.705$_{0.010}$ & 0.456$_{0.245}$ \\

        \bottomrule
    \end{tabular}
    }

    \resizebox{\textwidth}{!}{%
    \renewcommand\arraystretch{1.5}
    \begin{tabular}{lccccccccccccccccc}
        \toprule
         \multirow{2}*{\bf Method} & \multicolumn{7}{c}{\bf Image+Tabular} & \multicolumn{8}{c}{\bf Image+Text+Tabular} & \multirow{2}*{\bf avg $\uparrow$} & \multirow{2}*{\bf mrr $\uparrow$}\\

         \cmidrule(lr){2-8} \cmidrule(lr){9-16}

         & ccd & HAM & wikiart & cd18 & DVM & \textbf{avg} & \textbf{mrr} & petfinder & covid & artm & seattle & goodreads & KARD & \textbf{avg} & \textbf{mrr}  \\

         \midrule

         Baseline+ & 0.916$_{0.008}$ & 0.926$_{0.003}$ & 0.757$_{0.002}$ & 0.617$_{0.007}$ & 0.932$_{0.002}$ & 0.830$_{0.004}$ & 0.572$_{0.157}$ & 0.399$_{0.002}$ & 0.865$_{0.011}$ & 0.603$_{0.010}$ & 0.624$_{0.017}$ & 0.244$_{0.009}$ & 0.804$_{0.007}$ & 0.590$_{0.003}$ & 0.593$_{0.094}$  & 0.672$_{0.002}$ & 0.521$_{0.116}$ \\

        Input Aug. & 0.923$_{0.007}$ & 0.935$_{0.005}$ & 0.764$_{0.001}$ & 0.603$_{0.011}$ & 0.936$_{0.002}$ & \textbf{0.832$_{0.002}$} & \textbf{0.800$_{0.054}$} & 0.398$_{0.010}$ & 0.862$_{0.027}$ & 0.590$_{0.007}$ & 0.642$_{0.023}$ & 0.260$_{0.004}$ & 0.810$_{0.003}$ & \textbf{0.594$_{0.002}$} & \textbf{0.731$_{0.076}$}  & \textbf{0.676$_{0.001}$} & \textbf{0.653$_{0.018}$} \\

        Feature Aug.(Inde.) & 0.915$_{0.005}$ & 0.920$_{0.005}$ & 0.742$_{0.004}$ & 0.550$_{0.014}$ & 0.896$_{0.007}$ & 0.805$_{0.003}$ & 0.333$_{0.059}$ & 0.371$_{0.015}$ & 0.820$_{0.010}$ & 0.603$_{0.012}$ & 0.531$_{0.028}$ & 0.201$_{0.003}$ & 0.773$_{0.011}$ & 0.550$_{0.006}$ & 0.329$_{0.043}$ & 0.657$_{0.001}$ & 0.417$_{0.047}$ \\

        Feature Aug.(Joint) & 0.912$_{0.005}$ & 0.925$_{0.003}$ & 0.748$_{0.004}$ & 0.610$_{0.005}$ & 0.933$_{0.000}$ & 0.826$_{0.002}$ & 0.456$_{0.087}$ & 0.389$_{0.007}$ & 0.853$_{0.009}$ & 0.592$_{0.019}$ & 0.632$_{0.010}$ & 0.246$_{0.012}$ & 0.805$_{0.005}$ & 0.586$_{0.005}$ & 0.481$_{0.114}$ & 0.672$_{0.001}$ & 0.540$_{0.056}$ \\
        
        \bottomrule
    \end{tabular}
    }
    
    \label{tab:aug_detail}
\end{table}

\subsection{Handling Modality Missingness}
\begin{table}[H]\Huge
    \caption{Results of handling modality missingness on \textit{partial datasets with missingness}.}
    \centering
    \resizebox{\textwidth}{!}{%
    \renewcommand\arraystretch{1.5}
    \begin{tabular}{lcccccccccccccccc}
        \toprule
         \multirow{2}*{\bf Method} &\multicolumn{5}{c}{\bf Text+Tabular} & \multicolumn{4}{c}{\bf Image+Text} \\
         
         \cmidrule(lr){2-6} \cmidrule(lr){7-10} 
         
         &fake & airbnb & cloth  & \textbf{avg} & \textbf{mrr} & memotion &  fakeddit & \textbf{avg} & \textbf{mrr} \\
 
         % & roc\_auc & acc & acc & r2 & r2 & r2 &  \\
        \midrule
         Baseline+ & 0.956$_{0.009}$ & 0.417$_{0.002}$ & 0.752$_{0.006}$ & 0.708$_{0.003}$ & 0.546$_{0.182}$ & 0.586$_{0.004}$ & 0.879$_{0.002}$ & 0.732$_{0.003}$ & 0.583$_{0.297}$ \\
         
         Modality Dropout & 0.972$_{0.002}$ & 0.418$_{0.006}$ & 0.753$_{0.003}$ & 0.714$_{0.003}$ & 0.833$_{0.236}$ & 0.592$_{0.000}$ & 0.873$_{0.002}$ & 0.732$_{0.001}$ & 0.539$_{0.086}$ \\
       
         Learnable Embed(Numeric) & 0.956$_{0.009}$ & 0.416$_{0.005}$ & 0.752$_{0.006}$ & 0.708$_{0.005}$ & 0.467$_{0.126}$ & 0.586$_{0.004}$ & 0.879$_{0.002}$ & 0.732$_{0.003}$ & 0.583$_{0.297}$ \\
         
         Learnable Embed(Image) & 0.956$_{0.009}$ & 0.417$_{0.002}$ & 0.752$_{0.006}$ & 0.708$_{0.003}$ & 0.546$_{0.182}$ & 0.586$_{0.004}$ & 0.876$_{0.004}$ & 0.731$_{0.001}$ & 0.569$_{0.109}$ \\
         
        Modality Drop.+Learn. Embed(Image) & 0.972$_{0.002}$ & 0.418$_{0.006}$ & 0.753$_{0.003}$ & \textbf{0.714$_{0.003}$} & \textbf{0.833$_{0.236}$} & 0.591$_{0.001}$ & 0.880$_{0.004}$ & \textbf{0.736$_{0.002}$} & \textbf{0.681$_{0.255}$} \\
         
        \bottomrule
    \end{tabular}
    }

    \resizebox{\textwidth}{!}{%
    \renewcommand\arraystretch{1.5}
    \begin{tabular}{lcccccccccccccccccccccc}
        \toprule
         \multirow{2}*{\bf Method} &\multicolumn{5}{c}{\bf Image+Tabular} & \multicolumn{6}{c}{\bf Image+Text+Tabular} & \multirow{2}*{\bf avg$\uparrow$} & \multirow{2}*{\bf mrr$\uparrow$}\\
         
         \cmidrule(lr){2-6} \cmidrule(lr){7-12} 
         
         & HAM &  cd18 & DVM & \textbf{avg} & \textbf{mrr} & petfinder & covid & seattle & KARD & \textbf{avg} & \textbf{mrr}  \\

         \midrule

         Baseline+ & 0.926$_{0.003}$ & 0.617$_{0.007}$ & 0.932$_{0.002}$ & 0.825$_{0.004}$ & 0.593$_{0.069}$ & 0.399$_{0.002}$ & 0.865$_{0.011}$ & 0.624$_{0.017}$ & 0.804$_{0.007}$ & 0.673$_{0.001}$ & 0.514$_{0.103}$  & \textbf{0.730$_{0.002}$} & 0.560$_{0.059}$ \\

        Modality Dropout & 0.925$_{0.001}$ & 0.606$_{0.012}$ & 0.906$_{0.001}$ & 0.812$_{0.004}$ & 0.359$_{0.128}$ & 0.397$_{0.014}$ & 0.823$_{0.024}$ & 0.671$_{0.002}$ & 0.821$_{0.004}$ & 0.678$_{0.006}$ & 0.515$_{0.087}$  & \textbf{0.730$_{0.002}$} & 0.560$_{0.059}$ \\

        Learnable Embed(Numeric) & 0.922$_{0.004}$ & 0.617$_{0.007}$ & 0.932$_{0.002}$ & 0.824$_{0.002}$ & 0.531$_{0.137}$ & 0.399$_{0.002}$ & 0.850$_{0.008}$ & 0.630$_{0.008}$ & 0.804$_{0.007}$ & 0.671$_{0.002}$ & 0.436$_{0.059}$ & 0.728$_{0.002}$ & 0.492$_{0.052}$ \\

        Learnable Embed(Image) & 0.926$_{0.003}$ & 0.621$_{0.022}$ & 0.934$_{0.000}$ & \textbf{0.827$_{0.008}$} & \textbf{0.800$_{0.178}$} & 0.399$_{0.002}$ & 0.865$_{0.011}$ & 0.624$_{0.017}$ & 0.806$_{0.004}$ & 0.674$_{0.004}$ & 0.517$_{0.080}$  & \textbf{0.730$_{0.003}$} & \textbf{0.604$_{0.060}$} \\

        Modality Drop.+Learn. Embed(Image) & 0.921$_{0.006}$ & 0.594$_{0.032}$ & 0.906$_{0.002}$ & 0.807$_{0.012}$ & 0.265$_{0.068}$ & 0.393$_{0.008}$ & 0.829$_{0.023}$ & 0.676$_{0.007}$ & 0.820$_{0.004}$ & \textbf{0.679$_{0.006}$} & \textbf{0.603$_{0.030}$}  & 0.729$_{0.002}$ & 0.589$_{0.070}$\\
        
        \bottomrule
    \end{tabular}
    }
    \label{tab:missingness_detail}
\end{table}

We also investigate the effect of modality dropout trick on datasets without missingness, as it can also be applied to datasets without missing data. In \Cref{tab:modality_dropout}, modality dropout improves the model performance in text+tabular and image+text compositions.

\begin{table}[H]\Huge
    \caption{Results of handling modality missingness on \textit{partial datasets without missingness}.}
    \centering
    \resizebox{0.8\textwidth}{!}{%
    \renewcommand\arraystretch{1.5}
    \begin{tabular}{lcccccccccccccccccccc}
        \toprule
  
         \multirow{2}*{\bf Method} &\multicolumn{5}{c}{\bf Text+Tabular} & \multicolumn{5}{c}{\bf Image+Text} \\

         \cmidrule(lr){2-6} \cmidrule(lr){7-11}
         
         & channel & qaa & qaq  & \textbf{avg} & \textbf{mrr} & ptech & food101 & aep & \textbf{avg} & \textbf{mrr} \\
 
         % & roc\_auc & acc & acc & r2 & r2 & r2 &  \\
        \midrule
         Baseline+ & 0.511$_{0.004}$ & 0.480$_{0.004}$ & 0.483$_{0.009}$ & \textbf{0.491$_{0.005}$} & 0.667$_{0.136}$ & 0.556$_{0.006}$ & 0.928$_{0.002}$ & 0.557$_{0.005}$ & 0.681$_{0.002}$ & 0.556$_{0.079}$ \\
       
         Modality Dropout & 0.515$_{0.003}$ & 0.473$_{0.003}$ & 0.487$_{0.007}$ & \textbf{0.491$_{0.004}$} & \textbf{0.833$_{0.136}$} & 0.598$_{0.017}$ & 0.934$_{0.002}$ & 0.570$_{0.027}$ & \textbf{0.701$_{0.014}$} & \textbf{0.944$_{0.079}$}  \\

        \bottomrule
    \end{tabular}
    }

    \resizebox{0.8\textwidth}{!}{%
    \renewcommand\arraystretch{1.5}
    \begin{tabular}{lcccccccccccccccccccc}
        \toprule
  
         \multirow{2}*{\bf Method} &\multicolumn{4}{c}{\bf Image+Tabular} & \multicolumn{4}{c}{\bf Image+Text+Tabular} & \multirow{2}*{\bf avg$\uparrow$} & \multirow{2}*{\bf mrr$\uparrow$}\\

         \cmidrule(lr){2-5} \cmidrule(lr){6-9}
         
         & ccd & wikiart & \textbf{avg} & \textbf{mrr} & artm & goodreads & \textbf{avg} & \textbf{mrr}  \\
 
         % & roc\_auc & acc & acc & r2 & r2 & r2 &  \\
        \midrule
         Baseline+ & 0.916$_{0.008}$ & 0.757$_{0.002}$ & \textbf{0.836$_{0.005}$} & \textbf{0.833$_{0.118}$} & 0.603$_{0.010}$ & 0.244$_{0.009}$ & \textbf{0.423$_{0.010}$} & \textbf{0.833$_{0.118}$}  & 0.604$_{0.002}$ & 0.700$_{0.071}$ \\
       
         Modality Dropout &  0.920$_{0.002}$ & 0.736$_{0.003}$ & 0.828$_{0.001}$ & 0.667$_{0.118}$ & 0.599$_{0.024}$ & 0.227$_{0.013}$ & 0.413$_{0.014}$ & 0.667$_{0.118}$ & \textbf{0.606$_{0.007}$} &\textbf{0.800$_{0.071}$} \\

        \bottomrule
    \end{tabular}
    }
    \label{tab:modality_dropout}
\end{table}

\Cref{tab:missingness} examines the scenario where the missing ratios are similar in the training and test sets. To further explore the effect of the tricks in scenarios with varying missing modalities, especially with different missing ratios in the training and test sets, we use 10 datasets from our benchmark without missingness and randomly dropped modalities at missing ratios of 10\%, 30\%, and 50\%. The experiment results are shown in \Cref{tab:missingness_simulate}. Modality dropout generally enhances model performance in most scenarios. However, when the missing ratio in the training set is high, it can have a negative effect. We attribute this to the fact that a high training missing ratio, combined with further dropout, may lead to severe missingness, which can hinder effective model learning.
\begin{table}[H]\Huge
    \caption{Results of simulating modality missingness.}
    \centering
    \resizebox{\textwidth}{!}{%
    \renewcommand\arraystretch{1.5}
    \begin{tabular}{lll|cccccccccc}
        \toprule
         % \multirow{2}*{\bf Method.} & Tr. Miss. Rate & \multicolumn{5}{c}{\bf Text+Tabular} & \multicolumn{4}{c}{\bf Image+Text} & \multicolumn{5}{c}{\bf Image+Tabular} & \multicolumn{6}{c}{\bf Image+Text+Tabular} & \multirow{2}*{\bf avg$\uparrow$} & \multirow{2}*{\bf mrr$\uparrow$}\\
         % &fake & airbnb & cloth  & \textbf{avg} & \textbf{mrr} & memotion &  fakeddit & \textbf{avg} & \textbf{mrr} & HAM &  cd18 & DVM & \textbf{avg} & \textbf{mrr} & petfinder & covid & seattle & KARD & \textbf{avg} & \textbf{mrr}  \\

         \multirow{2}*{\bf Tr. Miss. Ratio} & \multirow{2}*{\bf Te. Miss. Ratio}  & \multirow{2}*{\bf Method.} & \multicolumn{2}{c}{\bf Text+Tabular} & \multicolumn{2}{c}{\bf Image+Text} & \multicolumn{2}{c}{\bf Image+Tabular} & \multicolumn{2}{c}{\bf Image+Text+Tabular} & \multirow{2}*{\bf avg$\uparrow$} & \multirow{2}*{\bf mrr$\uparrow$} \\
         &&& \textbf{avg} & \textbf{mrr} & \textbf{avg} & \textbf{mrr} &\textbf{avg} & \textbf{mrr} & \textbf{avg} & \textbf{mrr}  \\

         % & roc\_auc & acc & acc & r2 & r2 & r2 &  \\
        \midrule
        % train=0.1, test=0.1
        \multirow{12}*{\bf 0.1} & \multirow{4}*{\bf 0.1} & Baseline+ & 0.270 & 0.556  & 0.651 & 0.333 & 0.800 & 0.375 & 0.385 & \textbf{0.750}  & 0.513 & 0.492\\

         & & Modality Dropout &  \textbf{0.466} & \textbf{0.778}& \textbf{0.691} & \textbf{1.000} & 0.807 & \textbf{0.750} & 0.376 & 0.667  & \textbf{0.584} & \textbf{0.817}\\
         
          && Learnable Embed(Image) & 0.270 & 0.556  & 0.667 & 0.389  & 0.805 & 0.417 & \textbf{0.396} & 0.625 & 0.521 & 0.492\\
          
          && Modality Drop.+Learn. Embed(Image)  & \textbf{0.466} & \textbf{0.778} & 0.643 & 0.361  & \textbf{0.810} & \textbf{0.750}  & 0.372 & 0.625 & 0.569 & 0.617\\

          \cline{2-13}
          
         % train=0.1, test=0.3
        & \multirow{4}*{\bf 0.3} & Baseline+  & 0.243 & 0.556  & 0.576 & 0.361  & 0.738 & 0.250  & 0.333 & 0.333  & 0.459 & 0.392\\

         & & Modality Dropout & \textbf{0.407} & \textbf{0.778}  & \textbf{0.614} & \textbf{1.000}  & 0.758 & 0.417 & 0.334 & 0.417  & \textbf{0.525} & \textbf{0.700} \\
         
          && Learnable Embed(Image) &  0.243 & 0.556  & 0.581 & 0.389  & 0.757 & 0.667 & \textbf{0.351} & 0.667  & 0.469 & 0.550 \\
          
          && Modality Drop.+Learn. Embed(Image)  & \textbf{0.407} & \textbf{0.778} & 0.557 & 0.333  & \textbf{0.764} & \textbf{1.000} & 0.345 & \textbf{0.750} & 0.511 & 0.683\\

        \cline{2-13}

        % train=0.1, test=0.5
        & \multirow{4}*{\bf 0.5} & Baseline+ & 0.199 & 0.556 &  0.487 & 0.361 & 0.688 & 0.250 & 0.418 & 0.333  & 0.428 & 0.398 \\

         & & Modality Dropout & \textbf{0.327} & \textbf{0.778} &  \textbf{0.518} & \textbf{1.000}  & \textbf{0.714} & \textbf{1.000}  & \textbf{0.452} & \textbf{1.000} & \textbf{0.491} & \textbf{0.926}\\
         
          && Learnable Embed(Image) & 0.199 & 0.556 & 0.493 & 0.444 & 0.703 & 0.417  & 0.412 & 0.250 & 0.433 & 0.454 \\
          
          && Modality Drop.+Learn. Embed(Image)  & \textbf{0.327} & \textbf{0.778} & 0.467 & 0.333  & 0.708 & 0.417 & 0.435 & 0.500 & 0.470 & 0.519\\
         \cline{1-13}

        % train=0.3, test=0.1
        \multirow{12}*{\bf 0.3} & \multirow{4}*{\bf 0.1} & Baseline+  & \textbf{0.453} & \textbf{0.778}  & \textbf{0.657} & \textbf{0.667} & 0.782 & 0.292 & \textbf{0.384} & \textbf{0.750} & \textbf{0.566} & \textbf{0.642}\\

         & & Modality Dropout &   \textbf{0.453} & 0.556  & 0.653 & 0.778  & 0.788 & 0.500  & 0.375 & 0.667 & 0.564 & 0.633\\
         
          && Learnable Embed(Image) &  \textbf{0.453} & \textbf{0.778}  & 0.643 & 0.444 & 0.784 & 0.292  & 0.379 & 0.667& 0.561 & 0.558\\
          
          && Modality Drop.+Learn. Embed(Image)  & \textbf{0.453} & 0.556  & 0.610 & 0.250 & \textbf{0.796} & \textbf{1.000}  & 0.347 & 0.250 & 0.548 & 0.492\\
         \cline{2-13}

        % train=0.3, test=0.3
           & \multirow{4}*{\bf 0.3} & Baseline+ & 0.394 & 0.333  & \textbf{0.592} & \textbf{0.778} & 0.734 & 0.333 & 0.343 & 0.667  & 0.511 & 0.533\\

         & & Modality Dropout &   \textbf{0.403} & \textbf{1.000}  & 0.585 & 0.667  & 0.740 & 0.750  & \textbf{0.358} & \textbf{0.750} & \textbf{0.516} & \textbf{0.800} \\
         
          && Learnable Embed(Image) & 0.394 & 0.333 & 0.567 & 0.389 & 0.728 & 0.292  & 0.346 & 0.500 & 0.503 & 0.375\\
          
          && Modality Drop.+Learn. Embed(Image)  & \textbf{0.403} & \textbf{1.000}  & 0.537 & 0.250 & \textbf{0.746} & \textbf{1.000}  & 0.324 & 0.250 & 0.496 & 0.625 \\
          
         \cline{2-13}

        % train=0.3, test=0.5

           & \multirow{4}*{\bf 0.5} & Baseline+ & 0.317 & 0.333  & \textbf{0.498} & 0.611 & 0.686 & 0.417 & 0.424 & 0.250  & 0.471 & 0.435 \\

         & & Modality Dropout & \textbf{0.340} & \textbf{1.000}  & \textbf{0.498} &\textbf{ 0.833} & 0.682 & 0.417  & 0.463 & 0.500  & \textbf{0.482} & \textbf{0.759}\\
         
          && Learnable Embed(Image) & 0.317 & 0.333  & 0.474 & 0.389  & 0.674 & 0.250  & \textbf{0.486} & \textbf{1.000}  & 0.467 & 0.407\\
          
          && Modality Drop.+Learn. Embed(Image)  & \textbf{0.340} & \textbf{1.000} & 0.454 & 0.250  & \textbf{0.697} & \textbf{1.000} & 0.446 & 0.333  & 0.469 & 0.676\\
          
         \cline{1-13}

          % train=0.5, test=0.1
        \multirow{12}*{\bf 0.5} & \multirow{4}*{\bf 0.1} & Baseline+ & \textbf{0.443} & \textbf{0.778}  & \textbf{0.654} & \textbf{0.750}  & 0.772 & 0.375 &  0.336 & 0.417  & \textbf{0.551} & \textbf{0.617}\\

         & & Modality Dropout & 0.442 & 0.556 & 0.638 & 0.444  & 0.774 & 0.417 & \textbf{0.338} & \textbf{0.625}  & 0.546 & 0.508\\
         
          && Learnable Embed(Image) & 0.442 & 0.556  & 0.547 & 0.361  & 0.772 & 0.625& \textbf{0.338} & \textbf{0.625}  & 0.519 & 0.592 \\
          
          && Modality Drop.+Learn. Embed(Image)  & \textbf{0.443} & \textbf{0.778} & 0.609 & 0.528 & \textbf{0.782} & \textbf{0.750} & 0.336 & 0.500 & 0.539 & 0.575\\
          
         \cline{2-13}

          % train=0.5, test=0.3
        & \multirow{4}*{\bf 0.3} & Baseline+ &   0.389 & \textbf{0.778}  & \textbf{0.590 }& \textbf{0.778}  & 0.736 & 0.375  & \textbf{0.314} & \textbf{0.667}  & \textbf{0.504} & \textbf{0.675}\\

         & & Modality Dropout  & \textbf{0.390} & \textbf{0.778} & 0.577 & 0.417  & 0.736 & 0.667  & 0.306 & 0.375  & 0.499 & 0.567\\
         
          && Learnable Embed(Image) &   0.389 & \textbf{0.778} & 0.475 & 0.361 & 0.724 & 0.292 & 0.304 & 0.667 & 0.465 & 0.533\\
          
          && Modality Drop.+Learn. Embed(Image)  & \textbf{0.390} & \textbf{0.778} & 0.544 & 0.528 & \textbf{0.750} & \textbf{0.750}  & 0.292 & 0.417  & 0.489 & 0.625\\
         \cline{2-13}

          % train=0.5, test=0.5
       & \multirow{4}*{\bf 0.5} & Baseline+  & 0.313 & 0.556  & \textbf{0.490} & \textbf{0.611}  & 0.687 & 0.375  & \textbf{0.281} & 0.500  & 0.434 & 0.525\\

         & & Modality Dropout & \textbf{0.326} & \textbf{0.778}  & 0.487 & 0.583  & 0.680 & 0.667 & 0.279 & \textbf{0.625}  & \textbf{0.435} & \textbf{0.667}\\
         
          && Learnable Embed(Image) &  0.313 & 0.556  & 0.394 & 0.361 & 0.673 & 0.292  & 0.250 & \textbf{0.625} & 0.397 & 0.458\\
          
          && Modality Drop.+Learn. Embed(Image)  & \textbf{0.326} & \textbf{0.778}  & 0.461 & 0.528  & \textbf{0.700} & \textbf{0.750} & 0.259 & 0.333  & 0.428 & 0.608\\
         % \cline{2-13}

        \bottomrule
    \end{tabular}
    }
    
    \label{tab:missingness_simulate}
\end{table}

\subsection{Integrating Bag of Tricks}

\begin{table}[H]\Huge
    \caption{Results of integrating bag of tricks.}
    \centering
    \resizebox{\textwidth}{!}{%
    \renewcommand\arraystretch{1.5}
    \begin{tabular}{lccccccccccccccc}
        \toprule
         \multirow{2}*{\bf Method}  & \multicolumn{8}{c}{\bf Text+Tabular} & \multicolumn{7}{c}{\bf Image+Text} \\
         % & \multicolumn{7}{c}{\bf Image+Tabular} & \multicolumn{8}{c}{\bf Image+Text+Tabular} & \multirow{2}*{\bf avg $\uparrow$} & \multirow{2}*{\bf mrr $\uparrow$}

         \cmidrule(lr){2-9} \cmidrule(lr){10-16}  
         % \cmidrule(lr){17-23}  \cmidrule(lr){24-31} 

         & fake & airbnb & channel & qaa & qaq & cloth & \textbf{avg} & \textbf{mrr} & ptech & memotion & food101 & aep & fakeddit & \textbf{avg} & \textbf{mrr} \\
 
         % & roc\_auc & acc & acc & r2 & r2 & r2 &  \\
        \midrule
         Stacking & 0.968$_{0.004}$ & 0.42$_{0.003}$ & 0.516$_{0.000}$ & 0.454$_{0.011}$ & 0.485$_{0.024}$ & 0.753$_{0.003}$ & 0.599$_{0.004}$ & 0.519$_{0.052}$ & 0.705$_{0.002}$ & 0.591$_{0.000}$ & 0.934$_{0.000}$ & 0.719$_{0.024}$ & 0.900$_{0.006}$ & 0.770$_{0.004}$ & 0.589$_{0.057}$ \\

         Average All & 0.979$_{0.001}$ & 0.495$_{0.003}$ & 0.502$_{0.002}$ & 0.450$_{0.004}$ & 0.526$_{0.004}$ & 0.761$_{0.002}$ & 0.619$_{0.000}$ & 0.667$_{0.039}$ & 0.626$_{0.007}$ & 0.592$_{0.000}$ & 0.948$_{0.001}$ & 0.646$_{0.020}$ & 0.895$_{0.000}$ & 0.741$_{0.004}$ & 0.578$_{0.031}$ \\

         Ensemble Selection  & 0.974$_{0.003}$ & 0.498$_{0.002}$ & 0.511$_{0.006}$ & 0.447$_{0.004}$ & 0.528$_{0.004}$ & 0.762$_{0.000}$ & \textbf{0.620$_{0.001}$} & \textbf{0.676$_{0.086}$} & 0.732$_{0.018}$ & 0.589$_{0.003}$ & 0.949$_{0.000}$ & 0.722$_{0.013}$ & 0.914$_{0.002}$ & \textbf{0.781$_{0.004}$} & \textbf{0.889$_{0.096}$} \\
       
        \bottomrule
    \end{tabular}
    }

    \resizebox{\textwidth}{!}{%
    \renewcommand\arraystretch{1.5}
    \begin{tabular}{lccccccccccccccccc}
        \toprule
         \multirow{2}*{\bf Method} & \multicolumn{7}{c}{\bf Image+Tabular} & \multicolumn{8}{c}{\bf Image+Text+Tabular} & \multirow{2}*{\bf avg $\uparrow$} & \multirow{2}*{\bf mrr $\uparrow$}\\

         \cmidrule(lr){2-8} \cmidrule(lr){9-16}

         & ccd & HAM & wikiart & cd18 & DVM & \textbf{avg} & \textbf{mrr} & petfinder & covid & artm & seattle & goodreads & KARD & \textbf{avg} & \textbf{mrr}  \\
 
         % & roc\_auc & acc & acc & r2 & r2 & r2 &  \\
        \midrule
         Stacking & 0.947$_{0.005}$ & 0.924$_{0.006}$ & 0.797$_{0.003}$ & 0.689$_{0.003}$ & 0.966$_{0.001}$ & \textbf{0.865$_{0.002}$} & 0.600$_{0.072}$ & 0.394$_{0.012}$ & 0.916$_{0.004}$ & 0.618$_{0.011}$ & 0.706$_{0.007}$ & 0.213$_{0.008}$ & 0.808$_{0.004}$ & 0.609$_{0.005}$ & 0.426$_{0.035}$ & 0.701$_{0.002}$ & 0.528$_{0.023}$ \\

         Average All & 0.938$_{0.003}$ & 0.940$_{0.002}$ & 0.794$_{0.003}$ & 0.652$_{0.007}$ & 0.944$_{0.000}$ & 0.854$_{0.001}$ & 0.489$_{0.016}$ & 0.434$_{0.002}$ & 0.899$_{0.005}$ & 0.636$_{0.007}$ & 0.666$_{0.005}$ & 0.271$_{0.006}$ & 0.845$_{0.002}$ & 0.625$_{0.002}$ & 0.491$_{0.035}$ & 0.702$_{0.001}$ & 0.558$_{0.022}$ \\

         Ensemble Selection  &  0.950$_{0.002}$ & 0.935$_{0.007}$ & 0.809$_{0.005}$ & 0.661$_{0.011}$ & 0.968$_{0.000}$ & \textbf{0.865$_{0.004}$} & \textbf{0.800$_{0.082}$} & 0.428$_{0.007}$ & 0.937$_{0.006}$ & 0.655$_{0.008}$ & 0.712$_{0.001}$ & 0.294$_{0.001}$ & 0.884$_{0.000}$ & \textbf{0.652$_{0.003}$} & \textbf{0.917$_{0.000}$}  & \textbf{0.721$_{0.001}$} & \textbf{0.818$_{0.021}$} \\
       
        \bottomrule
    \end{tabular}
    }
    
    \label{tab:integrate_detail}
\end{table}

\subsection{Per Dataset Improvement Analysis}
In addition to analyzing the individual impacts of each trick, we explored performance from a per-task perspective. This helps identify which datasets are easy (where current performance is already near optimal) and which are more challenging (requiring further research to improve performance). Specifically, we calculated the performance gap between the baseline and ensemble selection, normalizing it by the gap between the baseline and the optimal value.

According to \Cref{tab:per_dataset_relative_improvement}, datasets such as airbnb, channel, qaa, qaq, memotion, petfinder, and artm show no or small improvements. Additionally, despite large improvements in datasets like cloth, ptech, aep, cd18, seattle, and goodreads, their best absolute performance remains below 0.8, indicating significant room for improvement. Conversely, datasets like fake, food101, ccd, HAM, DVM, and covid exhibit large relative improvements and high absolute performance (>0.9), suggesting less room for further enhancement.

\begin{table}[H]\Huge
    \caption{Per task relative improvement measured by (Ensemble Selection - Baseline) / (1 - Baseline).}
    \centering
    \resizebox{0.8\textwidth}{!}{%
    \renewcommand\arraystretch{1.5}
    \begin{tabular}{lccccccccccc}
        \toprule
         \multirow{2}*{}  & \multicolumn{6}{c}{\bf Text+Tabular} & \multicolumn{5}{c}{\bf Image+Text} \\
         % & \multicolumn{7}{c}{\bf Image+Tabular} & \multicolumn{8}{c}{\bf Image+Text+Tabular} & \multirow{2}*{\bf avg $\uparrow$} & \multirow{2}*{\bf mrr $\uparrow$}

         \cmidrule(lr){2-7} \cmidrule(lr){8-12}  
         % \cmidrule(lr){17-23}  \cmidrule(lr){24-31} 

         & fake & airbnb & channel & qaa & qaq & cloth & ptech & memotion & food101 & aep & fakeddit \\
 
         % & roc\_auc & acc & acc & r2 & r2 & r2 &  \\
        \midrule
         Baseline & 
         0.738 & 
         0.406 & 
         0.521 & 
         0.420 & 
         0.458 & 
         0.606 & 
         0.556 & 
         0.592 & 
         0.717 & 
         0.542 & 
         0.867 \\

         Ensemble Selection  & 0.974 & 
         0.498 & 
         0.511 & 
         0.447 & 
         0.528 & 
         0.762 &  
         0.732 & 
         0.589 & 
         0.949 & 
         0.722 & 
         0.914 \\

         Relative Improvement  & 0.901 & 
         0.155 & 
         -0.021 & 
         0.047 & 
         0.129 & 
         0.396 &
         0.396 &
         -0.007 & 
         0.82 & 
         0.393 & 
         0.353 \\
       
        \bottomrule
    \end{tabular}
    }

    \resizebox{0.8\textwidth}{!}{%
    \renewcommand\arraystretch{1.5}
    \begin{tabular}{lccccccccccc}
        \toprule
         \multirow{2}*{} & \multicolumn{5}{c}{\bf Image+Tabular} & \multicolumn{6}{c}{\bf Image+Text+Tabular} \\

         \cmidrule(lr){2-6} \cmidrule(lr){7-12}

         & ccd & HAM & wikiart & cd18 & DVM & petfinder & covid & artm & seattle & goodreads & KARD \\
 
         % & roc\_auc & acc & acc & r2 & r2 & r2 &  \\
        \midrule
         Baseline & 
         0.914 & 
         0.913 & 
         0.740 & 
         0.539 & 
         0.917 & 
         0.353 & 
         0.829 & 
         0.608 & 
         0.478 & 
         0.050 & 
         0.770 \\

         Ensemble Selection  &  0.950 & 
         0.935 & 
         0.809 & 
         0.661 & 
         0.968 & 
         0.428 & 
         0.937 & 
         0.655 & 
         0.712 & 
         0.294 & 
         0.884 \\

         Relative Improvement  &
         0.419  &
         0.253  &
         0.265  &
         0.265  &
         0.614  &
         0.116  &
         0.632  &
         0.12  &
         0.448  &
         0.257  &
         0.496\\
       
        \bottomrule
    \end{tabular}
    }
    
    \label{tab:per_dataset_relative_improvement}
\end{table}

\subsection{Training Throughput Analysis}
Another key aspect of analyzing the tricks is computational latency. We analyze the training throughput for different tricks. The experiments were conducted using a single A10 GPU with a batch size of 1 across four datasets (Cloth, Fakeddit, DVM, and KARD), representing four different modality combinations. Key findings from \Cref{tab:training_througput} include:
\begin{itemize}
    \item Basic tricks don’t increase training overhead.
    \item LF-Transformer is slower than LF-MLP due to the larger transformer fusion module.
    \item LF-LLM is the slowest, attributed to its large model size.
    \item LF-Aligned is faster than Baseline+ due to its smaller model size.
    \item Early-fusion is the fastest, benefiting from a single unified encoder.
    \item Input and independent feature augmentation don’t slow down training.
    \item Modality dropout and learnable embeddings maintain baseline training speed.
    \item Ensemble selection is the slowest, as it requires training models with individual tricks.
\end{itemize}

\begin{table}[H]
    \caption{Training throughput (samples / s) comparison of different tricks. The experiments were conducted using a single A10 GPU with a batch size of 1 across four datasets (Cloth, Fakeddit, DVM, and KARD), representing 4 different modality combinations of image, text, and tabular data.}
    \centering
    \resizebox{\textwidth}{!}{%
    \renewcommand\arraystretch{1.5}
    \begin{tabular}{l|cccc}
        \toprule
         {\bf Tricks} & {\bf Text+Tabular} &  {\bf Image+Text } &  {\bf Image+Tabular } &  {\bf Image+Text+Tabular}\\
         & (cloth) & (fakeddit) & (DVM) & (KARD) \\
         \midrule
Baseline (LF-MLP)  & 13.69 & 7.59 & 11.93 & 7.1 \\
+Greedy Soup & 13.66 & 7.6 & 11.91 & 7.11 \\
+Graident Clip & 13.67 & 7.72 & 12.15 & 7.09 \\
+ LR Warmup & 14.08 & 7.65 & 11.53 & 7.13 \\
+LR Decay (Baseline+) & 14.31 & 7.53 & 12.05 & 7.12 \\
LF-Transformer & 12.83 & 7.35 & 11.17 & 6.73 \\
LF-Aligned & 21.03 & 10.65 & 11.91 & 9.53 \\
LF-LLM & 1.81 & 1.65 & 1.8 & 1.55 \\
LF-SF & 14.55 & 7.69 & 12.09 & 7.38 \\
Early-fusion & 22.79 & 23.68 & 22.35 & 22.53 \\
Convert Categorical & 14.75 & 7.53 & 7.57 & 6.88 \\
Convert Numeric & 13.02 & 7.53 & 10.67 & 7.23 \\
Positive-only & 14.09 & 7.63 & 11.52 & 7.15 \\
Positive+Negative & 10.39 & 5.65 & 9.17 & 5.41 \\
Input Aug. & 14.11 & 7.61 & 11.89 & 7.13 \\
Feature Aug.(Inde.) & 14.13 & 7.46 & 11.72 & 7.06 \\
Feature Aug.(Joint) & 13.04 & 7.31 & 10.93 & 6.78 \\
Modality Dropout & 14.26 & 7.71 & 11.95 & 7.19 \\
Learnable Embed(Numeric) & 14.28 & 7.86 & 11.97 & 7.28 \\
Learnable Embed(Image) & 14.12 & 7.51 & 11.95 & 7.09 \\
Modality Drop.+Learn. Embed(Image) & 14.16 & 7.89 & 11.59 & 7.2 \\
Ensemble Selection & 0.664 & 0.387 & 0.544 & 0.362 \\

        \bottomrule
    \end{tabular}
    }

    \label{tab:training_througput}
\end{table}

\subsection{Tabular Importance Analysis}
 To evaluate the usefulness of the tabular modality in the benchmark, we conducted experiments excluding tabular data, using the same setup as Baseline+. The performance gap between Baseline+ with and without tabular data indicates the added predictive value of tabular data for each dataset. As shown in \Cref{tab:tabular_importance_analysis}, tabular data significantly enhance predictive performance in datasets such as channel, cd18, DVM, covid, seattle, and KARD, but contribute less in datasets like fake, qaa, qaq, cloth, ccd, HAM, and goodreads. This highlights the diversity of our benchmark in terms of the predictive power of tabular data. Note that datasets containing only image and text data do not include tabular data, so their performance remains unchanged.

\begin{table}[H]\Huge
    \caption{Performance comparison w/ and w/o tabular data.}
    \centering
    \resizebox{\textwidth}{!}{%
    \renewcommand\arraystretch{1.5}
    \begin{tabular}{lccccccccccc}
        \toprule
         \multirow{2}*{}  & \multicolumn{6}{c}{\bf Text+Tabular} & \multicolumn{5}{c}{\bf Image+Text} \\
         % & \multicolumn{7}{c}{\bf Image+Tabular} & \multicolumn{8}{c}{\bf Image+Text+Tabular} & \multirow{2}*{\bf avg $\uparrow$} & \multirow{2}*{\bf mrr $\uparrow$}

         \cmidrule(lr){2-7} \cmidrule(lr){8-12}  
         % \cmidrule(lr){17-23}  \cmidrule(lr){24-31} 

         & fake & airbnb & channel & qaa & qaq & cloth & ptech & memotion & food101 & aep & fakeddit \\
 
         % & roc\_auc & acc & acc & r2 & r2 & r2 &  \\
        \midrule
        Baseline+ & 
        0.956$_{0.009}$ & 
        {\bf 0.417$_{0.002}$} & {\bf 0.511$_{0.004}$} & {\bf 0.480$_{0.004}$} & {\bf 0.483$_{0.009}$} & 0.752$_{0.006}$ & 0.556$_{0.006}$ & 0.586$_{0.004}$ & 0.928$_{0.002}$ & 0.557$_{0.005}$ & 0.879$_{0.002}$ \\
        
         Baseline+ (w/o tabular) & 
         {\bf 0.972$_{0.010}$} & 
         0.379$_{0.009}$ & 
         0.322$_{0.006}$ & 
         0.478$_{0.003}$ & 
		0.476$_{0.011}$ & 
          0.752$_{0.007}$ & 
          0.556$_{0.006}$ & 0.586$_{0.004}$ & 0.928$_{0.002}$ & 0.557$_{0.005}$ & 0.879$_{0.002}$ \\

        \bottomrule
    \end{tabular}
    }

    \resizebox{\textwidth}{!}{%
    \renewcommand\arraystretch{1.5}
    \begin{tabular}{lccccccccccc}
        \toprule
         \multirow{2}*{} & \multicolumn{5}{c}{\bf Image+Tabular} & \multicolumn{6}{c}{\bf Image+Text+Tabular} \\

         \cmidrule(lr){2-6} \cmidrule(lr){7-12}

         & ccd & HAM & wikiart & cd18 & DVM & petfinder & covid & artm & seattle & goodreads & KARD \\
 
         % & roc\_auc & acc & acc & r2 & r2 & r2 &  \\
        \midrule
         Baseline+ & 
         0.916$_{0.008}$ & 
         {\bf 0.926$_{0.003}$} & {\bf 0.757$_{0.002}$} & {\bf 0.617$_{0.007}$} & {\bf 0.932$_{0.002}$} & {\bf 0.399$_{0.002}$} & {\bf 0.865$_{0.011}$} & {\bf 0.603$_{0.010}$} & {\bf 0.624$_{0.017}$} & {\bf 0.244$_{0.009}$} & {\bf 0.804$_{0.007}$} \\

         Baseline+ (w/o tabular)  &  {\bf 0.926$_{0.007}$} & 
         0.918$_{0.003}$ & 
         0.729$_{0.004}$ & 
         0.339$_{0.022}$ & 
         0.71$_{0.002}$ & 
         0.344$_{0.012}$ & 
         0.779$_{0.041}$ & 
         0.588$_{0.015}$ & 
         0.508$_{0.016}$ & 
         0.239$_{0.020}$ & 
         0.641$_{0.013}$ \\

        \bottomrule
    \end{tabular}
    }
    
    \label{tab:tabular_importance_analysis}
\end{table}

\subsection{Comparison to Strong Tabular Baselines}
We compared our bag of tricks with two strong tabular baselines: weighted ensembling and stacking+bagging. Within the ensembling/stacking framework, we employed several popular tabular models: XGBoost, CatBoost, LightGBM, Extremely Randomized Trees, and MLP. These models were trained using 5-fold cross-validation (bagging), and 2-layer stacking was applied. According to \Cref{tab:tabular_baselines_comparison}, our proposed bag-of-tricks outperforms tabular stacking in most datasets due to the inclusion of additional image/text information and effective multimodal tricks. However, tabular stacking achieves better performance on three datasets: channel, cd18, and seattle, underscoring the importance of tabular data. Performance could potentially be further improved by ensembling both these tabular models and our bag of tricks. Nevertheless, the primary focus of this paper is on exploring multimodal tricks rather than unimodal ones.

\begin{table}[H]\Huge
    \caption{Performance comparison with strong tabular baselines. Within the tabular ensembling/stacking framework, we employed several popular tabular models: XGBoost, CatBoost, LightGBM, Extremely Randomized Trees, and MLP.}
    \centering
    \resizebox{0.8\textwidth}{!}{%
    \renewcommand\arraystretch{1.5}
    \begin{tabular}{lccccccccccc}
        \toprule
         \multirow{2}*{}  & \multicolumn{6}{c}{\bf Text+Tabular} & \multicolumn{5}{c}{\bf Image+Text} \\
         % & \multicolumn{7}{c}{\bf Image+Tabular} & \multicolumn{8}{c}{\bf Image+Text+Tabular} & \multirow{2}*{\bf avg $\uparrow$} & \multirow{2}*{\bf mrr $\uparrow$}

         \cmidrule(lr){2-7} \cmidrule(lr){8-12}  
         % \cmidrule(lr){17-23}  \cmidrule(lr){24-31} 

         & fake & airbnb & channel & qaa & qaq & cloth & ptech & memotion & food101 & aep & fakeddit \\
 
         % & roc\_auc & acc & acc & r2 & r2 & r2 &  \\
        \midrule
        Weighted Ensemble (tabular) & 0.718 & 0.43 & 0.545 & 0.076 & 0.041 & -0.005 & - & - & - & - & - \\
        Stacking+Bagging (tabular) & 0.726 & 0.451 & {\bf 0.55} & 0.076 & 0.045 & 0.003 & - & - & - & - & - \\

         Ensemble Selection (ours)  & {\bf 0.974} & {\bf 0.498} & 0.511 & {\bf 0.447} & {\bf 0.528} & {\bf 0.762} &  0.732 & 0.589 & 0.949 & 0.722 & 0.914 \\

        \bottomrule
    \end{tabular}
    }

    \resizebox{0.8\textwidth}{!}{%
    \renewcommand\arraystretch{1.5}
    \begin{tabular}{lccccccccccc}
        \toprule
         \multirow{2}*{} & \multicolumn{5}{c}{\bf Image+Tabular} & \multicolumn{6}{c}{\bf Image+Text+Tabular} \\

         \cmidrule(lr){2-6} \cmidrule(lr){7-12}

         & ccd & HAM & wikiart & cd18 & DVM & petfinder & covid & artm & seattle & goodreads & KARD \\
 
         % & roc\_auc & acc & acc & r2 & r2 & r2 &  \\
        \midrule
         Weighted Ensemble (tabular) & 
         0.557 & 0.734 & 0.548 & 0.72 & 0.951 & 0.345 & 0.883 & 0.554 & 0.719 & 0.018 & 0.869 \\

         Stacking+Bagging (tabular) & 
         0.557 & 0.739 & 0.545 & {\bf 0.737} & 0.952 & 0.361 & 0.892 & 0.571 & {\bf 0.729} & 0.02 & 0.87 \\

         Ensemble Selection (ours)  &  {\bf 0.950} & {\bf 0.935} & {\bf 0.809} & 0.661 & {\bf 0.968} & {\bf 0.428} & {\bf 0.937} & {\bf 0.655} & 0.712 & {\bf 0.294} & {\bf 0.884} \\

        \bottomrule
    \end{tabular}
    }
    
    \label{tab:tabular_baselines_comparison}
\end{table}

\end{document}